\documentclass{article}

\usepackage[nonatbib, final]{neurips_2022}

\usepackage{placeins}
\usepackage{float}
\usepackage{multirow}

\usepackage{amsmath}
\usepackage{amssymb,amsfonts,accents}

\usepackage{algorithmic}
\usepackage{graphicx}

\usepackage{textcomp}
\usepackage{bbm}
\usepackage{comment}
\usepackage{dsfont}
\usepackage{caption}
\usepackage{subcaption}
\usepackage{tabularx}
\usepackage{slashbox}
\usepackage[flushleft]{threeparttable}
\usepackage{multirow}

\usepackage{comment}
\usepackage{svg}
\usepackage{tcolorbox}
\usepackage{booktabs}

\usepackage{empheq}

\usepackage{amsthm}

\newtheorem{theorem}{Theorem}[section]
\newtheorem{lemma}[theorem]{Lemma}

\newcolumntype{C}[1]{>{\centering\arraybackslash}p{#1}}

\usepackage{xcolor}
\newcommand\bcolorbox[2]{\bcbaux{#1}#2 \endbcb}
\def\bcbaux#1#2 #3\endbcb{%
  \colorbox{#1}{\strut#2}%
  \ifx\relax#3\relax\def\next{}\else%
    \colorbox{#1}{ \strut}%
    \allowbreak%
    \def\next{\bcbaux{#1}#3\endbcb}%
  \fi%
  \next%
}
\newcommand{\best}[1]{\bcolorbox{black!30}{#1}}

\newcommand*{\err}[1]{\scalebox{0.8}{$ \pm#1$}}%

\newcommand\norm[1]{\lVert#1\rVert}

\newcounter{equationset}

\usepackage{tikz}
\newcommand*\circled[1]{\tikz[baseline=(char.base)]{\node[shape=circle,draw,inner sep=0.01pt] (char) {#1};}}

\usepackage[page, header]{appendix}

\usepackage{wrapfig}
\usepackage{wrapfig, blindtext}
 
\title{Learning Mixture Structure on Multi-Source Time Series for Probabilistic Forecasting}

\author{
  Tian Guo \\ 
  Systematic Equities Team \\
  RAM Active Investments \\
  Geneva, Switzerland \\
  \texttt{tig@ram-ai.com} \\
}

\begin{document}

\maketitle

\begin{abstract}
In many data-driven applications, collecting data from different sources is increasingly desirable for enhancing performance.
In this paper, we are interested in the problem of probabilistic forecasting with multi-source time series.
We propose a neural mixture structure-based probability model for learning different predictive relations and their adaptive combinations from multi-source time series.
We present the prediction and uncertainty quantification methods that apply to different distributions of target variables.
Additionally, given the imbalanced and unstable behaviors observed during the direct training of the proposed mixture model, we develop a phased learning method and provide a theoretical analysis.
In experimental evaluations, the mixture model trained by the phased learning exhibits competitive performance on both point and probabilistic prediction metrics.
Meanwhile, the proposed uncertainty conditioned error suggests the potential of the mixture model's uncertainty score as a reliability indicator of predictions. 
\end{abstract}

\section{Introduction}

Time series data are prevalent in many applications~\cite{lin2017hybrid, lim2021time}, and in this paper, we are interested in probabilistic forecasting in the setting of multi-source time series.
The data source can be distinct locations, entities, records, sensors, etc., from where data is collected. 
This multi-source data setting can fit into a variety of scenarios, where learning from a single source may be sub-optimal.
For instance, in finance, different operations by market participants (e.g., performing transactions, putting limit orders, etc.) are separately recorded, thereby giving rise to distinct sources of features reflecting the underlying characteristics of markets~\cite{Ns2006, stosic2018collective}.
In environmental monitoring where sensors are usually deployed in multiple sites, each site is a data source, and collectively using the data from different sensor sites would be favorable to predictive analytics~\cite{wu2016data}.

\begin{wrapfigure}{R}{0.6\textwidth}
  \begin{center}
    \includegraphics[width=0.6\textwidth]{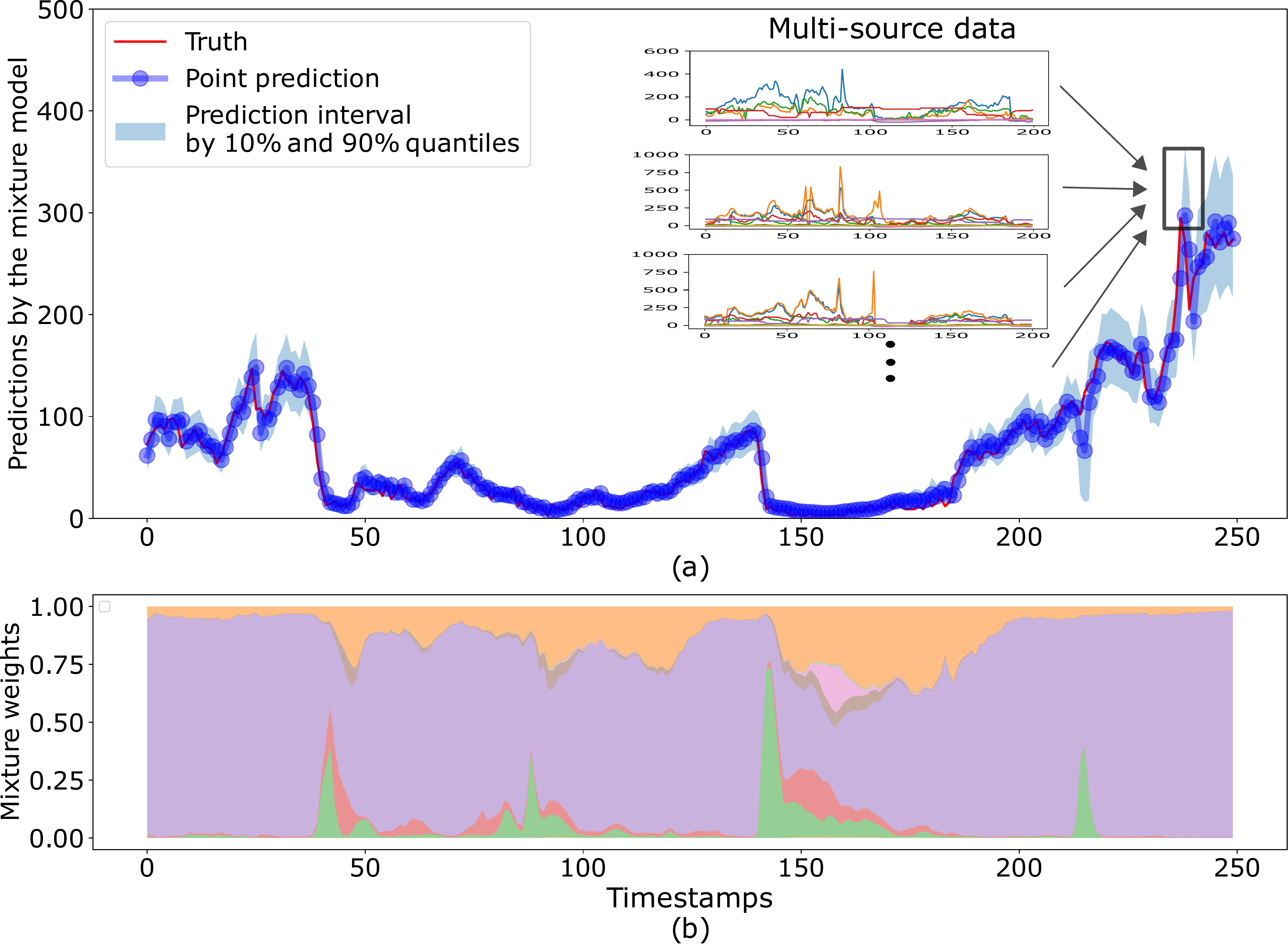}
  \end{center}
  \caption{
  (a) Illustration of multi-source time series based probabilistic forecasting. 
      The prediction at each timestep (e.g., highlighted by the rectangle) is based on the multi-source data in a look-back window.  
  (b) Time-varying mixture weights of data sources in the prediction.
      Each color corresponds to a data source.
      At each timestamp, the larger the colored area, the higher the weight and the more relevance to the target the corresponding data source has.
      The sum of all colored value ranges is $1.0$ at each timestamp.
  }\label{fig:intro_multi_src_pred}
\end{wrapfigure}

\textbf{Challenges.}
Multi-source time series data usually carry different dynamics information and have time-varying relevance to the target variable~\cite{gamboa2017deep, zhu2018spatiotemporal}, as illustrated in Fig.~\ref{fig:intro_multi_src_pred}.
Meanwhile, the target variable might follow different distributions depending on applications~\cite{salinas2020deepar}.
For downstream decision-making, besides point predictions, it is increasingly preferred to provide additional predictive information, e.g., uncertainty score, quantiles, etc~\cite{wang2019deep, gasthaus2019probabilistic, abdar2021review}.
All of the above highlight the need for a flexible and adaptive probabilistic model structure on multi-source data.

As for the probability mixture model, it is typically composed of hierarchical components, and the interactions between these components during the training affect the end performance.
For instance, as shown in Sec.~\ref{sec:issues}, direct optimizing our mixture model's loss function leads to imbalanced learning behaviors of different data-source-specific components, implying that some data sources' predictive power is inadequately or even biasedly captured.
This further hinders exploiting the full capacity of mixture models.

\textbf{Contributions.} Specifically, the contribution is as follows:

(1) We present the neural mixture structure based probability model on multi-source time series. 
It is equipped with a representation module to encode multi-source data, and a representation sharing mechanism to serve the prediction and weight modules.

(2) We present the adaptive prediction and uncertainty quantification methods and analyze the implication of different components in the uncertainty score.
This inference process is applicable to different target distributions, e.g., normal, and log-normal distributions in this paper.

(3) We demonstrate the imbalanced learning behaviors when directly minimizing the proposed mixture model's loss function, and then develop the phased learning method with theoretical analysis.

(4) Through experiments on real datasets from finance and environmental monitoring scenarios, our mixture model trained by phased learning exhibits competitive performance on both point and probabilistic prediction metrics.
Meanwhile, we put forward uncertainty conditioned error analysis to compare uncertainty scores from different methods.

\section{Related work} \label{sec:related}

\textbf{Probabilistic Forecasting on Time Series.}
Probabilistic forecasting focuses on the predictive probabilistic characteristics of target variables and is aimed to provide a variety of predictive quantities for downstream decision-making, e.g., uncertainty scores, predictive quantiles, prediction intervals, etc~\cite{gasthaus2019probabilistic, salinas2020deepar, alaa2020frequentist}.

Recently, many neural network based probabilistic forecasting models have been proposed for time series data.
\cite{rangapuram2018deep, wang2019deep} respectively equipped RNNs with the linear state space models and random effect components.
\cite{salinas2020deepar} used RNNs with autoregressive and covariate inputs for probabilistic forecasts.
\cite{rasul2020multivariate, rasul2021autoregressive} focused on modeling multivariate time series.
Stochastic sequential models \cite{chung2015recurrent, fraccaro2016sequential, goyal2017z, krishnan2017structured, pal2021rnn, qiu2021history} developed stochastic temporal latent states for modeling generative dynamics in time series.
Given the setting of multi-source data in this paper, the above methods mostly model the input data as a whole and could hardly differentiate various time-varying relations in multi-source data, thereby potentially leading to inferior predictive performance.

\textbf{Mixture Models.}
Thanks to the flexibility and adaptivity, mixture models are applied in various areas, such as unsupervised learning~\cite{eigen2013learning, shi2019variational, kurle2019multi}, natural language processing~\cite{shen2019mixture, cho2019mixture, du2021glam}, computer vision~\cite{li2017infogail, varamesh2020mixture, vowels2021vdsm}, interpretable machine learning~\cite{guo2019exploring}, etc.
Few attempts have been made to explore mixture model structures, and learning methods for probabilistic forecasting with multi-source data.

As a category of latent variable models, mixture models are normally trained through (stochastic) expectation maximization (EM) style optimizations~\cite{jain2017non}.
\cite{balakrishnan2017statistical} provided theoretical convergence analysis for linear mixture models.
\cite{chen2018stochastic} integrated variance reduction into stochastic EM algorithms.
In this paper, we explore the practical treatments for stably and balancedly learning different data-source-corresponding modules of the mixture model.

\section{Problem and Proposed Model}\label{sec:model}
\subsection{Problem Statement}
Let $y_t$ be the value of the target variable at time $t$.
Assume $S$ number of different data sources, which are indexed by $s = 1, \cdots, S$. 
The multi-source data used to predict $y_t$ is denoted by $\mathbf{x}_{1:S,<t} = \{\mathbf{x}_{s,<t}\}_{s=1}^{S}$, where each element $\mathbf{x}_{s,<t}$ is a multi-dimensional time series from a limited window of historical data of source $s$ prior to timestamp $t$.
Depending on applications, the latest timestamp in $\mathbf{x}_{s,<t}$ could be $t-1$ for one-step or $t-h$ for $h$-step ahead prediction.
For simplicity, we skip the notations for time series dimensions and history window length.

The task is to model the predictive density function $p(y_t \,|\, \mathbf{x}_{1:S,<t})$ for providing predictive quantities such as point predictions, uncertainty scores and so on~\cite{mackay2003information, yuksel2012twenty}.
Note that in this paper, we assume that $\mathbf{x}_{1:S,<t}$ implicitly contains the auto-regressive data of the target variable, i.e., $y_{<t}$ when available, either as an individual data source or being included into one.

\subsection{Model Structure}\label{sec:model_structure}
The idea of the mixture structure on $p(y_t \,|\, \mathbf{x}_{1:S,<t})$ is to introduce a latent discrete random variable $z_t$ for differentiating various predictive relations between $\mathbf{x}_{1:S,<t}$ and $y_t$.
In our case, $z_t$ is defined on the data source index set, i.e., $z_t \in \{1, \cdots, S\}$, thereby leading to data-source-wise prediction modules.
The probability mass function of $z_t$ serves as the weight module.

Meanwhile, instead of using the raw data $\mathbf{x}_{1:S,<t}$ as inputs to the weight and prediction modules, we feed a set of latent representations that are learned from $\mathbf{x}_{1:S,<t}$ by powerful neural networks.
Let $\mathbf{h}_{1:S,t}$ denote this set of representation vectors $\{ \mathbf{h}_{s,t} \}_{s=1}^{S}$.
The factorization of the joint density function including $z_t$ and $\mathbf{h}_{1:S,t}$ as latent variables is expressed as: 
\begin{align}\label{eq:share}
\begin{split}
p(y_t \,|\, \mathbf{x}_{1:S,<t}) = \int \underbrace{p( \, \mathbf{h}_{1:S,t} \,|\, \mathbf{x}_{1:S,<t} )}_{\text{representation module}}
\cdot 
\sum_{s = 1}^{S} \underbrace{\mathbb{P}(z_t = s \,|\, \mathbf{h}_{1:S,t} )}_{\text{weight module}}
\cdot 
\underbrace{p(y_t \,|\, z_t = s, \mathbf{h}_{s,t})}_{\text{prediction module}} \, 
\text{d}\mathbf{h}_{1:S,t}
\end{split}
\end{align}

\textit{Representation Module} is to encode each source's data into the latent representations that will serve the prediction and weight modules.
There are various probabilistic representation learning methods for time series, e.g., state-space models, stochastic recurrent neural networks, etc~\cite{doerr2018probabilistic, franceschi2019unsupervised, qiu2021history}.
Since advancing the representation learning is not the focus of this paper, we choose a simple yet flexible way to formulate the representation module, namely, placing a Dirac delta distribution centered at the hidden state output of a temporal data encoder~\cite{bayer2014learning, fraccaro2016sequential, karl2016deep}.
This module is formulated as:
\begin{align}
\begin{split}
p( \, \mathbf{h}_{1:S, t} \,|\, \mathbf{x}_{1:S,<t} ) = \prod_{s=1}^S \delta(\mathbf{h}_{s,t} - \widetilde{\mathbf{h}}_{s,t}) \,\,\,\,\, \text{s.t.} \, \widetilde{\mathbf{h}}_{s,t} = \text{NN}_{\eta_s}(\mathbf{x}_{s, <t}), \forall s = 1, \cdots, S
\end{split}
\end{align}
, where $\delta(a) = 1$ only when $a=0$.
$\text{NN}_{\eta_s}(\mathbf{x}_{s, <t})$ is the encoder to model the temporal dynamics of source $s$ and $\eta_s$ denotes the trainable parameter in it.
In the experiment, we employ an LSTM as the temporal encoder, while it is flexible to use others.

If it is hypothesized to there exist global latent factors governing the dynamics across data sources, the above formulation is extensible by learning a global representation from all sources and then sharing it across individual representations~\cite{li2018disentangled, wang2019deep}.

\textit{Prediction Module} is to parameterize the source-wise predictive distribution $p_{\omega_s}(y_t \,|\, z_t = s, \mathbf{h}_{s,t})$ with the learnable parameters $\omega_s$. 
For instance, for $y_t \in \mathbb{R}$ and the normal distribution is applied, we have $y_t \,| \mathbf{h}_{s,t} \sim \mathcal{N}(\mu_{s,t}, \sigma_{s,t}^2) $, where $\mu_{s,t}, \sigma_{s,t}^2 = g_{\omega_s}(\mathbf{h}_{s,t})$, and $g_{w_s}(\mathbf{h}_{s,t})$ can be any flexible function such as multiple dense layers parameterized by $\omega_s$~\cite{salinas2020deepar}.
For $y_t \in \mathbb{R}^{+}$, we can choose the log-normal distribution, i.e., $\log y_t \,| \mathbf{h}_{s,t} \sim \mathcal{N}(\mu_{s,t}, \sigma_{s,t}^2)$~\cite{cohen1980estimation, mackay2003information}.

\textit{Weight Module} is the probability mass function of $z_t$:
\begin{align}\label{eq:softmax_s}
\mathbb{P}_{\theta}(z_t = s \,|\, \mathbf{h}_{1:S,t} ) & 
 \triangleq 
\frac{\exp(\, f_{s}(\mathbf{h}_{s,t}) \,)}{ \sum_{k=1}^S\exp(\, f_{k}(\mathbf{h}_{k,t}) \,)}
\end{align}
, where $\theta_{s}$ denotes the parameters of the logit function for data source $s$.
$f_{\theta_s}( \cdot )$ can be any flexible function such as neural networks.
The trainable parameters from all logit functions constitute the parameter set of this module, i.e., $\theta = \{\theta_{s}\}_{s=1}^{S}$.

\section{Adaptive Inference and Phased Learning}\label{sec:learn}

\subsection{Inference}\label{sec:infer}
We mainly present the adaptive point prediction and uncertainty quantification. 
Other predictions like quantiles, intervals, etc., and their derivations are described in Appendix~\ref{appendix:infer}.
For simplicity, we express the weight module as $\mathbb{P}_{\theta}(z_{t} = s \,|\, \cdot )$.

\textbf{Point Prediction.}
It is the predictive mean of the mixture, which is the weighted sum of each source's predictive mean.
\begin{align}\label{eq:mean}
\hat{y}_t & \triangleq \mathbb{E}[y_t \,|\, \mathbf{x}_{1:S, <t} ] = \sum_{s=1}^S \mathbb{P}_{\theta}(z_{t} = s \,|\, \cdot ) \mathbb{E}[{y}_{t} | z_t=s, \widetilde{\mathbf{h}}_{s,t}]
\end{align}
, where $\mathbb{E}[{y}_{t} \,|\, z_t=s, \widetilde{\mathbf{h}}_s ]$ denotes the predictive mean by source $s$. 
For instance, it is $\mu_{s,t}$ for normal distribution and $\exp(\mu_{s,t} + \frac{1}{2}\sigma_{s,t}^2)$ for the log-normal distribution~\cite{mackay2003information}.

\textbf{Uncertainty Score.}
In forecasting tasks, the predictive variance is commonly used as an uncertainty measure~\cite{pearce2018high, qiu2020quantifying}.
With a single forward pass, our mixture model provides the uncertainty score $\hat{u}_t$ consisting of two components as follows:
\begin{align}\label{eq:var}
\begin{split}
\hat{u}_t = \underbrace{ \sum_{s = 1}^S \mathbb{P}_{\theta}(z_t = s | \cdot) \text{Var}(y_t | z_t = s, \widetilde{\mathbf{h}}_{s,t}) }_{\text{Aleatoric Uncertainty}}
& + \underbrace{ \sum_{s = 1}^S \mathbb{P}_{\theta}(z_t = s | \cdot) \, \mathbb{E}^2[y_t | z_t=s, \widetilde{\mathbf{h}}_{s,t} ] - \mathbb{E}^{2}\left[y_t \,|\, \mathbf{x}_{1:S, <t} \right]}_{\text{Mixture Uncertainty}}
\end{split}
\end{align}

\begin{wrapfigure}{R}{0.45\textwidth}
  \begin{center}
    \includegraphics[width=0.45\textwidth]{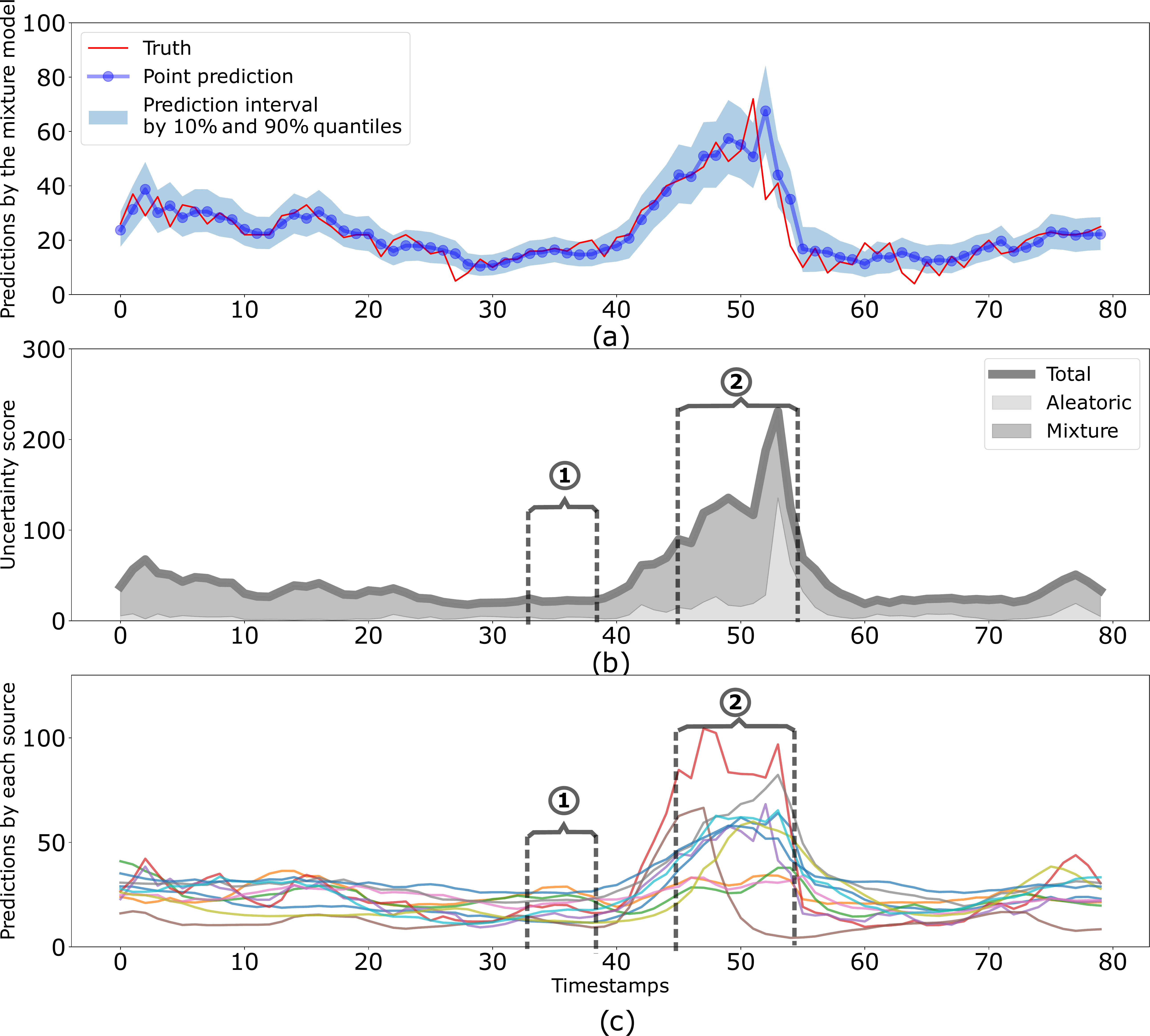}
  \end{center}
  \caption{
   Illustration of the predictive uncertainty by the mixture model on AIR data.
   (a) True values, point, and interval predictions.
   (b) Uncertainty and its component values. 
   The total uncertainty score shown by the thick line is the sum of values represented by the light and dark areas. 
   (c) Point predictions by data-source-specific prediction modules.
  }\label{fig:intro_uncertainty_pred}
\end{wrapfigure}

For instance, the variance term $\text{Var}(y_t | z_t = s, \widetilde{\mathbf{h}}_{s,t})$ is $\sigma_{s,t}^2$ for normal distribution, and $\exp(\sigma_{s,t}^2 -1) \cdot \exp(2 \mu_{s,t} + \sigma_{s,t}^2)$ for log-normal distribution~\cite{crow1987lognormal}.

In Eq.\ref{eq:var}, the aleatoric uncertainty quantifies predictive noise inherent in the target, while the mixture part reflects the variation among individual data sources' predictions.
The mixture weights serve to adjust the contributions of different sources. 
This is different from some uncertainty quantification methods using equal-weighted components, e.g., model snapshots, and latent states~\cite{gal2016dropout, lakshminarayanan2017simple, qiu2021history}.



In the following, through the testing examples in Fig.\ref{fig:intro_uncertainty_pred}, we give a qualitative interpretation of these two uncertainty components.
In time period \circled{1}, the target variable stays relatively stable in Fig.\ref{fig:intro_uncertainty_pred}(a).
Accordingly, in Fig.\ref{fig:intro_uncertainty_pred}(b), the aleatoric uncertainty is low.
In Fig.\ref{fig:intro_uncertainty_pred}(c), the predictions by individual sources show some consensus and reside in a close value range, and thus the mixture uncertainty is also low in \circled{1}.
On the other hand, in period \circled{2}, when the target is fluctuating in Fig.\ref{fig:intro_uncertainty_pred}(a), besides the aleatoric part, the mixture uncertainty drives up the total uncertainty because of the widened prediction dispersion as shown in Fig.\ref{fig:intro_uncertainty_pred}(c). 
Interestingly, Fig.\ref{fig:intro_uncertainty_pred}(c) also reveals that when the target is in a volatile period (e.g., \circled{2}), just some of the data sources can provide seasonably reasonable predictions, while the rest of the sources yield too high or too low predictions and seem to lack enough predictive power.






\subsection{Issues with Direct Learning}\label{sec:issues}
In this part, we analyze the imbalanced learning behaviors when directly minimizing the mixture loss. 
We first define the training dataset $\mathcal{D} = \big\{ y_t, \mathbf{x}_{1:S,<t} \big\}_{t \in \mathcal{T}}$, where $\mathcal{T}$ represents the set of timestamps at which data instances are collected.
Let $\Theta$ represent the set of trainable parameters, i.e., $\Theta = \left\{ \{\eta_s, \omega_s\}_{s=1}^S, \theta \right\}$.
Then, the mixture loss is expressed as:
\begin{align}\label{eq:loss}
\begin{split}
& \mathcal{L}\left( \Theta \,; \mathcal{D} \right) \triangleq -\frac{1}{|\mathcal{T}|} \sum_{t \in \mathcal{T}}
\log 
\mathbb{E}_{\mathbf{h}_{1:S,t}} 
\left[
\sum_{s = 1}^{S} \mathbb{P}_{\theta}(z_t = s | {\mathbf{h}}_{s,t} ) \, p_{{\omega}_{s}}(y_t | z_t = s, {\mathbf{h}}_{s,t}) 
\right]
\end{split}
\end{align}

\begin{wrapfigure}{R}{0.65\textwidth}
  \begin{center}
    \includegraphics[width=0.63\textwidth]{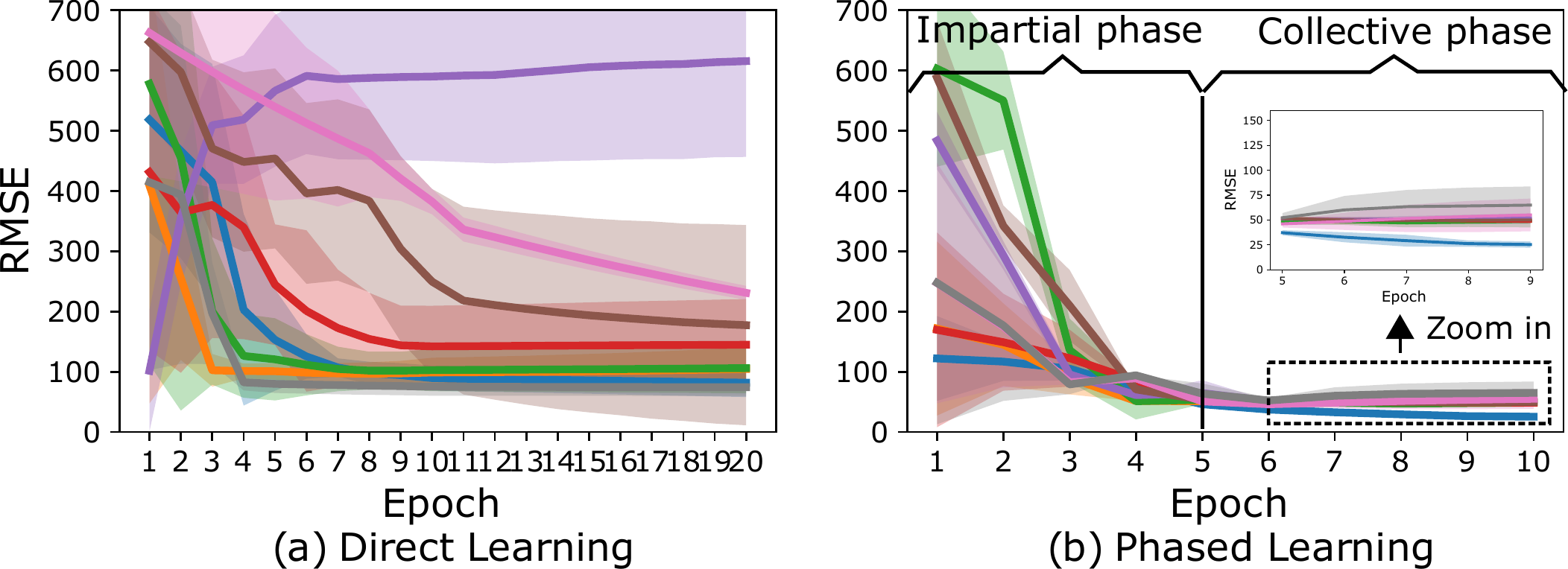}
  \end{center}
  \caption{
  Illustration of imbalanced and unstable behaviors of directly learning the mixture model, compared with the proposed phased learning.
  Each curve represents the rooted mean squared error (RMSE) between target values and predictive means by a source's prediction module over training epochs.
  The results are based on five different training runs on AIR data in Sec.\ref{sec:exp}.
  }\label{fig:tr_behaviour}
\end{wrapfigure}

Next, through Fig.\ref{fig:tr_behaviour}, we illustrate the imbalanced learning status of different data sources when directly minimizing the mixture loss via stochastic gradient descent (SGD) based methods~\cite{goodfellow2016deep}.
This imbalanced and unstable behavior might hinder the mixture model from adequately capturing different data sources' predictive relations and their combinations, and thus impairs the overall performance as shown in Sec.\ref{sec:exp}.

Specifically, Fig.\ref{fig:tr_behaviour}(a) reports the source-wise prediction error which is defined by the rooted mean squared error between training true values and predictive means by each source's prediction module over training epochs.
It shows that some error curves converge poorly, e.g., the purple one implies the corresponding source's prediction module hardly captures useful predictive relations.
As for the converged curves, the noticeable error dispersion suggests that those prediction modules might be learned to different extents.

As a comparison, Fig.\ref{fig:tr_behaviour}(b) shows the source-wise prediction error of the model trained by the proposed phased learning in Sec.\ref{sec:tr}.
The source-wise errors stably and rapidly decline to a close range, and, likely, all prediction modules are reasonably learned.
In the zoom-in part, the spread between converged error curves still implies the different predictive powers of data sources. 
Notably, in Fig.\ref{fig:tr_behaviour}(b), the purple error curve also converges nicely. 
It implies the corresponding data source has the predictive power, but it is not properly captured in Fig.\ref{fig:tr_behaviour}(a).
In Sec.\ref{sec:exp}, we will show that the overall prediction performance of the mixture model considerably benefits from this phased learning.


\begin{lemma}\label{lamma:grad}
In minimizing the mixture loss via SGD optimization, given a random training sample indexed by $t$, the vectorized derivative of a source's prediction and representation modules are denoted by:
\begin{align}
& \mathbf{g}_{\omega_{s}, t} \triangleq -\frac{\partial \log p_{\mathbf{\omega}_{s}}(y_t \,|\, z_t = s, \widetilde{\mathbf{h}}_{s,t})}{\partial \omega_{s}} \Big|_{\Theta}  \label{eq:grad_pred_content} 
\\
&\mathbf{g}_{\eta_{s}, t} \triangleq -\frac{\partial \log p_{\omega_{s}}(y_t \,|\, z_t = s, \widetilde{\mathbf{h}}_{s,t})}{\partial \widetilde{\mathbf{h}}_{s,t}} \nabla\widetilde{\mathbf{h}}_{s,t} \Big|_{\Theta}  \label{eq:grad_rep_content}
\\
& \mathbf{d}_{\eta_{s}, t} \triangleq -\sum_{k=1}^{S} \pi_{k,t} \frac{\partial \log\mathbb{P}_{\theta}(z_{t} = k | \cdot)}{\partial \widetilde{\mathbf{h}}_{s,t}} \nabla\widetilde{\mathbf{h}}_{s,t} \Big|_{\Theta} 
\end{align}
where $\nabla\widetilde{\mathbf{h}}_{s,t}$ denotes the Jacobian matrix of source $s$'s representation w.r.t. $\eta_s$.
Meanwhile, the posterior weight $\pi_{s, t}$ is defined as:
\begin{align}\label{eq:post_weight_content}
\begin{split}
\pi_{s,t} \triangleq p(z_t = s \,|\, y_t, \mathbf{x}_{1:S,<t} ; \Theta) = \frac{p_{{\omega}_{s}}(y_t \,|\, z_t = s, \widetilde{\mathbf{h}}_{s,t}) \, \mathbb{P}_{\theta}(z_t = s \,|\, \cdot)}{\sum_{k = 1}^{S} p_{{\omega}_{k}}(y_t \,|\, z_t = k, \widetilde{\mathbf{h}}_{k,t}) \, \mathbb{P}_{\theta}(z_t = k \,|\, \cdot)}  
\end{split}
\end{align}
Then, the (stochastic) updates to the prediction and representation parameters of source $s$, i.e., $\omega_{s}$ and $\eta_{s}$, are:
\vspace{-0.2cm}
\begin{align}\label{eq:update_content}
\pi_{s,t} \, \mathbf{g}_{\omega_{s},t} \,\,\,\text{and}\text{ }\,\,\, \pi_{s,t} \, \mathbf{g}_{\eta_{s},t} + \mathbf{d}_{\eta_{s},t}
\end{align}
(Derivation details in the appendix section.)
\end{lemma}

Lemma~\ref{lamma:grad} sheds light on one potential cause of the imbalanced learning, i.e., the intertwined relation between posterior weights and parameter updates, and Sec.\ref{sec:tr} will present another theoretical analysis. 
Specifically, at the beginning of the training, all representation, prediction, and weight modules are barely trained.
Some data sources receive low posterior weights (i.e., $\pi_{s,t}$ ), if they happen to have low (stochastic) predictive density values and/or low mixture weights, according to Eq.\ref{eq:post_weight_content}.
Then, based on Eq.\ref{eq:update_content}, low posterior weights lead to under-weighted (stochastic) updates and might slow down or even corrupt the learning progress of the corresponding source's modules. 
In turn, this posterior weight bias might persist and result in the imbalanced learning status in Fig.\ref{fig:tr_behaviour}(a). 

\subsection{Phased Learning}\label{sec:tr}
To relax the above issues, we propose the phased learning method splitting the training process into two phases respectively for unbiasedly learning each data source's predictive power and then the adaptive combinations.
(The proofs in this part are in the appendix section.)

\begin{lemma}\label{lemma:bound_loss}
In the iterative learning process, at any point of $\Theta$, there exists the upper bound of the mixture loss:
\begin{align}
\begin{split}
\mathcal{L}\left( \Theta \,; D_t\right) 
& \leq 
\overline{\mathcal{L}}\big( \{\eta_s, \omega_s\}_{s=1}^S \,; D_t \big)
\end{split}
\end{align}
where $D_t$ is a random training sample $D_t \triangleq ( y_t, \mathbf{x}_{1:S,<t} ) \in \mathcal{D}$.
\begin{align}\label{eq:bound_content}
\begin{split}
\overline{\mathcal{L}}\big( \{\eta_s, \omega_s\}_{s=1}^S  \,;  D_t \big) \triangleq - \sum_{s = 1}^{S} \frac{1}{S} \log  p_{\omega_s}(y_t \,|\, z_t = s, \widetilde{\mathbf{h}}_{s,t} ) - \log S a_{t}^{*}
\end{split}
\end{align}
and $a_{t}^{*}$ is a constant value, $a_{t}^{*} = \min_{s}(\, \{ \, \mathbb{P}(z_t = s \,|\, \cdot) \} \,)$. 
\end{lemma}


First, the \textbf{impartial phase} learning is inspired by Lemma~\ref{lemma:bound_loss}.
It minimizes the upper bound Eq.\ref{eq:bound_content}, thereby leading to the following update scheme:
\begin{equation}\label{eq:update_first}
\frac{1}{S}\mathbf{g}_{\omega_{s}, t}  \,\,\,\text{and}\text{ }\,\,\,  \frac{1}{S}\mathbf{g}_{\eta_{s}, t}
\end{equation}
, where $\mathbf{g}_{\omega_{s}, t} $ and $\mathbf{g}_{\eta_{s}, t}$ are defined in Eq.\ref{eq:grad_pred_content} and \ref{eq:grad_rep_content}.
These updates only involve the prediction-related parameters, i.e., the representation and prediction modules.
Meanwhile, the equal-weighted manner allows for impartially learning the predictive relation of each data source.

\begin{theorem}\label{theory:loss_decline}
For all stochastic derivatives $\mathbf{g}_{} \in \{ \mathbf{g}_{\omega_{s}}, \mathbf{g}_{\eta_{s}}, \mathbf{d}_{\eta_{s}} \}$ corresponding to the data source $s$, suppose it is bounded $\mathbb{E}\lVert \mathbf{g}_{} \rVert^2 \leq G^2$.
The trainable parameters related to a data source's prediction are denoted by 
$\Theta_{s} = [ \omega_{s}, \eta_{s} ]^\top$ and the corresponding gradient is  $\mathcal{G}_s = \mathbb{E}[\mathbf{g}_{\omega_{s}}, \mathbf{g}_{\eta_{s}}]^{\top}$.
Over the learning step $i = 0, \cdots, I$, the convergence of $\mathcal{G}_s$ to a stationary point in the direct optimization and the impartial-phase optimization are respectively: 
\begin{align}
\begin{split}
\text{Direct:} \,\, & \frac{1}{I} \sum_{i=0}^{I-1}  \mathbb{E} \norm{\mathcal{G}_{s,i}}^2 \leq \sqrt{10} G \sqrt{ \frac{ L \left( \mathcal{L}_{s,0} - \mathcal{L}_{s,I} \right) }{I \pi_{s}^{*}} }  + \sum_{i=0}^{I-1} \frac{1}{2I} \left( \left\lVert \mathbb{E}[ \mathbf{g}_{\eta_s,i} ] - \mathbb{E}[ \mathbf{d}_{\eta_s,i} ] \right\rVert^2 \right)
\end{split} \label{eq:converge_direct_content}
\\
\text{Impartial phase:} \,\, & \frac{1}{I} \sum_{i=0}^{I-1} \mathbb{E} \norm{\mathcal{G}_{s,i}}^2 
                               \leq 
                               2 G \sqrt{ \frac{ L \left(\mathcal{L}_{s, 0} - \mathcal{L}_{s, I} \right) }{I}}
                               \label{eq:converge_phase_content}
\end{align}
, where $\pi_{s}^{*} \in (0, 1)$ is the minimum of poster weights across data samples and iteration steps.
$\mathcal{L}_{s, i}$ represents the negative log-likelihood of the predictions by data source $s$ at step $i$.
\end{theorem}

Theorem~\ref{theory:loss_decline} explains the convergence patterns of the direct and impartial-phase learning observed in Fig.\ref{fig:tr_behaviour}.
As shown in Eq.\ref{eq:converge_direct_content}, there are two more factors that impact the convergence of each data source's prediction related parameters in direct learning, compared with impartial-phase learning.
The value of $\pi^*_s$ is less than $1$ and implies the slow value decline of the first term on the right-hand-side of Eq.\ref{eq:converge_direct_content}. 
Meanwhile, if the second term on the right-hand-side of Eq.\ref{eq:converge_direct_content} does not vanish completely, it converges to the neighborhood of a stationary point, i.e., the corresponding prediction modules might be inadequately trained.
For instance, the converged error curves in Fig.\ref{fig:tr_behaviour}(a) are still higher than the counterparts in Fig.\ref{fig:tr_behaviour}(b).


Second, the \textbf{collective phase} will update all modules' parameters by switching to minimize the original mixture loss Eq.\ref{eq:loss} until the end of the training.
The weigh module is now activated to learn the adaptive combinations.
The representations and prediction modules are updated as Eq.\ref{eq:update_content}.
Thanks to the reasonably learned representation and prediction modules in the {impartial phase}, the posterior weight now in Eq.\ref{eq:post_weight_content} more reliably reflects different data sources' relevance to the target.
In this phase, it amounts to fine-tuning the representations and prediction modules by considering the data source relevance reflected in posterior weights.
For instance, in Fig.\ref{fig:tr_behaviour}(b), the zoom-in figure shows that some sources' error curves continue to decline in the {collective phase}.

In practice, the number of epochs of the {impartial phase} can be tuned as a hyperparameter.
On the experiment datasets, the range of $5$ to $15$ epochs is enough for the {impartial phase}.
Meanwhile, we found that the total number of epochs needed in phased learning is comparable to the direct learning, i.e., having two phases does not incur the need for more epochs in total. 
Compared with the direct learning with a certain number of epochs, the phased learning can mostly present an enhanced performance by running the {impartial} and {collective phase} with the same budget of epochs.

\section{Experiments}\label{sec:exp}
In this section, we report the experimental comparison of different models and learning methods.
The Appendix contains additional experiment details and results.

\subsection{Baselines}
\textbf{DAR} denotes DeepAR, which uses auto-regressive recurrent neural networks to model the probabilistic distribution of the target variable~\cite{salinas2020deepar}.
\textbf{DF} refers to the Deep Factor model, which is a global-local method based on a global neural network backbone and local probabilistic random effect models~\cite{wang2019deep}.
\textbf{AF} represents Autoformer~\cite{wu2021autoformer}, a decomposition based Transformer~\cite{vaswani2017attention} variant with an auto-correlation mechanism.

DAR and DF are probability models, and AF is adapted for probabilistic forecasting based on~\cite{li2019enhancing};
hence all baselines can parameterize different distributions for the target variable and be evaluated on point and probabilistic prediction metrics.
For baselines, multi-source data are fed as a whole.
For the mixture model and baselines, probabilistic forecasts like predictive quantiles and intervals are derived via the same type of methods presented in \ref{appendix:infer}.

\subsection{Datasets}
Existing time series forecasting models mostly take as input single (univariate or multivariate) time series, while our mixture model works on multi (multivariate) time series from distinct sources.
Thus, we use the following data from two scenarios.
In each dataset, data instances are time-ordered, and we use the first $70\%$ of points for training, the next $10\%$ for validation, and the last $20\%$ for testing.

\textbf{AIR} is from the environmental monitoring scenario~\cite{zhang2017cautionary}.
There are twelve air-quality monitoring sites, and each site collects an eleven-dimensional time series including hourly air pollutant and meteorological observations, from March 2013 to February 2017.
The target variable, which is the air pollutant value reflected in PM2.5 measurements, lies in the non-negative domain, and thus the log-normal distribution is applied.
The time series from all sites forms twelve-source data for forecasting the target variable of each site.

\textbf{VOL} is from the finance area, i.e., two liquid cryptocurrency exchanges, Bitfinex and Bitstamp.
From the limit order book (LOB) and transaction records of each exchange, it collects two minute-level multi-dimensional feature time series, from May 2018 to September 2018.
The target variable is the intra-day traded volume of bitcoins.
It is the raw volume deduced by the intra-day periodicity and lies in the real value domain, and thus the normal distribution is applied.
In a cryptocurrency exchange, it records different trading-related operations, e.g., transaction data stores the information about executed trades, the limited order book (LOB) records the standing orders with buy or sell limits, etc.
We formulate multi-source data consisting of the time series from LOB and transaction features as well as historical volume.

\begin{table*}[t]
\caption{
Results of point and probabilistic prediction metrics (mean $\pm$ standard error). 
The best result is marked by the grey box. 
Due to the space limitation, the results of the rest of the sites in AIR data are in Appendix~\ref{appendix:exp}. 
}
\centering
\resizebox{1.0\textwidth}{!}{
\begin{tabular}{ll|llll|llll}
\toprule 
                     &        & \multicolumn{4}{c|}{RMSE $\downarrow$}                       & \multicolumn{4}{c}{MAE $\downarrow$} \\
                     & Data   & DAR & DF & AF & \textbf{MIX}             & DAR & DF & AF & \textbf{MIX}  \\
                      \midrule
\multirow{4}{*}{AIR} & Site 0 & 23.650 \err{0.645} & 23.792 \err{0.756} & 68.293 \err{0.985} & \best{19.022} \err{0.054}  
                              & 12.015 \err{0.223} & 13.743 \err{0.291} & 35.196 \err{0.307} & \best{10.297} \err{0.048}  \\
                     \cmidrule{2-10}
                     & Site 1 & 20.012 \err{0.205} & 31.467 \err{0.570} & 66.199 \err{0.912} & \best{18.517} \err{0.086} 
                              & 10.596 \err{0.109} & 13.247 \err{0.772} & 39.196 \err{0.307} & \best{9.929}  \err{0.027}  \\
                     \midrule
\multirow{4}{*}{VOL} & Exc. 0 & 1.206\err{0.00342} & 1.157\err{0.00389} & 1.892\err{0.00253}   & \best{1.149}\err{0.00168}     
                              & 0.925\err{0.00276} & 0.916\err{0.00259} & 1.021\err{0.00324}   & \best{0.898}\err{0.00588}  \\
                     \cmidrule{2-10}
                     & Exc. 1 & 1.382\err{0.0075}  & 1.271\err{0.00177} & 2.010\err{0.0098}    & \best{1.265}\err{0.00302}   
                              & 0.995\err{0.00464} & 0.981\err{0.00101} & 1.032\err{0.00433}   & \best{0.977}\err{0.00197}   \\
\bottomrule
\toprule 
                     &        & \multicolumn{4}{c|}{NLLm $\downarrow$}                              & \multicolumn{4}{c}{QLm $\downarrow$} \\
                     & Data   & DAR & DF & AF & \textbf{MIX}                  & DAR & DF & AF & \textbf{MIX} \\
                      \midrule
\multirow{4}{*}{AIR} & Site 0 & 3.922\err{0.011} & 4.196\err{0.013} &  5.439\err{0.004} & \best{3.879}\err{0.011}  
                              & 0.298\err{0.004} & 0.327\err{0.003} &  0.233\err{0.001} & \best{0.0463}\err{0.001}  \\
                     \cmidrule{2-10}
                     & Site 1 & 3.860\err{0.021} & 3.992\err{0.029} &  5.208\err{0.002} & \best{3.696}\err{0.003} 
                              & 0.231\err{0.004} & 0.202\err{0.004} &  0.237\err{0.001} & \best{0.0505}\err{0.001}  \\
                     \midrule
\multirow{4}{*}{VOL} & Exc. 0 & 1.594\err{0.00485} & 1.578\err{0.00550} & 1.601\err{0.00324}   & \best{1.541}\err{0.00258}       
                              & 2.974\err{0.0413}  & 4.498\err{0.0694}  & 4.982\err{ 0.0983}   & \best{2.840}\err{0.0624} \\
                     \cmidrule{2-10}
                     & Exc. 1 & 1.698\err{0.00853} & 1.660\err{0.00131} & 1.723\err{0.00732}   & \best{1.647}\err{0.00220}       
                              & 2.606\err{0.0958}  & 4.801\err{0.119}   & 5.012\err{0.1012}    & \best{2.353}\err{0.0930}  \\
\bottomrule
\end{tabular}
}
\label{tab:baselines}
\end{table*}

\subsection{Setup}
\textbf{Hyper-Parameters.}
The hyper-parameter search and associated training processes are run on a server with NVIDIA A100 GPUs.
For all models, Bayesian optimization is applied to search in the hyper-parameter space~\cite{bergstra2013making}.
Once the best hyperparameters are fixed, each model is retrained for five times with different random seeds, and the average performance is reported. 
The hyper-parameter space is detailed in Appendix.

\textbf{Point Prediction Metrics.}
The root-mean-square error (RMSE) and mean absolute error (MAE) are defined on true values and predictive means.

\textbf{Probabilistic Prediction Metrics.}
The mean negative log-likelihood (NLLm) is the predictive negative log-likelihood of testing instances averaged by the number of instances.
The mean quantile loss (QLm) is the mean of quantile losses w.r.t. a set of quantile levels~\cite{gasthaus2019probabilistic}.
A quantile loss is defined on true values and predictive quantiles to reflect how well the predictive distribution fits the target variable's probabilistic characterises~\cite{wang2019deep}.
Detailed formations of these metrics are in Appendix.

\subsection{Results}

\textbf{Comparison of Prediction Performance.}
In Table~\ref{tab:baselines}, on each site or exchange (exc.), the mixture model outperforms baselines on the point and probabilistic prediction metrics.
It indicates that the mixture models more accurately characterize the distributional properties of the target variable.
Compared with the baselines modeling multi-source data as a whole, the outperformance of the mixture model suggests that capturing adaptive relations in multi-source data can better harness the predictive power.




\textbf{Comparison of Training Methods.}
Table~\ref{tab:train_method_perf} shows the results of the mixture models respectively trained by different learning methods.
The phased learning noticeably improves the performance of the mixture model on all metrics, while in some cases the mixture model trained by the direct learning under-performs the baselines in Table~\ref{tab:baselines}.
This observation suggests that the specialized training process is beneficial for exploiting the capacity of mixture models. 
Meanwhile, the low standard errors of the phased learning imply the training process is relatively more stable, thereby leading to more consistent generalization performance. 
\FloatBarrier
\begin{table}[htbp]
\centering
\caption{
Performance of the mixture models respectively trained by the direct and phased learning methods (mean $\pm$ standard error).
Additional results are in Appendix~\ref{appendix:exp}. 
}
\resizebox{1.0\textwidth}{!}{
      \begin{tabular}{ll|ll|ll|ll|ll}
      \toprule      
&   & \multicolumn{2}{c|}{RMSE $\downarrow$} & \multicolumn{2}{c|}{MAE $\downarrow$} & \multicolumn{2}{c|}{NLLm $\downarrow$} & \multicolumn{2}{c}{QLm $\downarrow$} \\
& Data  & Direct & Phased    & Direct & Phased   & Direct & Phased   & Direct & Phased       \\
       \midrule
\multirow{4}{*}{AIR} & Site 0  & 27.173\err{1.617} & \best{19.022}\err{0.054}  & 14.905\err{0.423} & \best{10.297}\err{0.048}   
                               & 4.103\err{0.133}  & \best{3.879}\err{0.011}   & 0.0582\err{0.009} & \best{0.0463}\err{0.001} \\
                     \cmidrule{2-10}
                     & Site 1  & 41.549\err{2.013} & \best{18.517}\err{0.086}   & 17.876\err{1.102} &\best{9.929}\err{0.027}      
                               & 3.772\err{0.006}  & \best{3.696}\err{0.003}    & 0.0589\err{0.009} & \best{0.0505}\err{0.001}\\
                    \bottomrule        
                    \toprule
\multirow{4}{*}{VOL} & Exc. A  & 1.172\err{0.00123}   & \best{1.149}\err{0.00168}    & 0.934\err{0.0356}  & \best{0.898}\err{0.0588}     
                               & 1.596\err{0.00216}   & \best{1.541}\err{0.00258}    & 3.012\err{0.0563}  & \best{2.840}\err{0.0624} \\
                     \cmidrule{2-10}
                     & Exc. B  & 1.282\err{0.00203}   & \best{1.265}\err{0.00302}    & 0.998\err{0.00123}   & \best{0.977}\err{0.00197}    
                               & 1.656\err{0.00158}   & \best{1.647}\err{0.00220}    & 2.510\err{0.0890}  & \best{2.353}\err{0.0930} \\
      \bottomrule
      \end{tabular}
}
\label{tab:train_method_perf}
\end{table}
\FloatBarrier

\begin{figure}[htbp]
\centering
    \includegraphics[width=\textwidth]{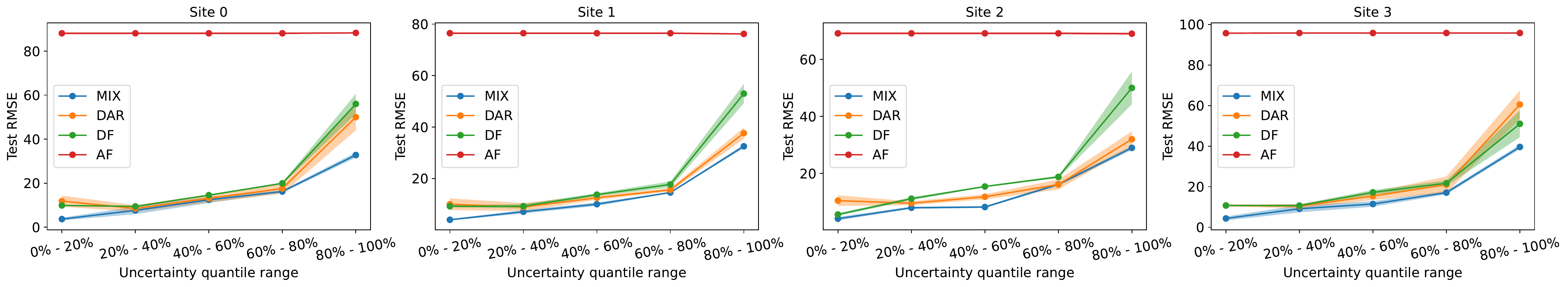}
\caption{Uncertainty Conditioned Prediction Errors.}
\label{fig:unc_error}
\end{figure}

\textbf{Comparison of Uncertainty Conditioned Errors.}
The uncertainty score can be an indicator of the prediction error level, namely, high uncertainty scores imply unreliable predictions and thus potentially large prediction errors, and vice versa.
Following this intuition, we propose the uncertainty-conditioned prediction error to compare the uncertainty scores from different models.
Specifically, given the predictions and associated uncertainty scores, we first derive the empirical quantiles of the uncertainty scores.
Then, as shown in Fig.\ref{fig:unc_error}, we calculate the prediction errors (RMSE) by using the predictions with the uncertainty scores falling within each interval of quantile levels.

In Fig.\ref{fig:unc_error}, based on the above intuition, the uncertainty-conditioned error curve is expected to present a monotonic increasing pattern.
The mixture model has monotonic error curves, while the baselines DAR and DF have some irregular errors, i.e., the error of the high uncertainty interval is lower than that of the low uncertainty interval.
Compared with the baselines only capturing the aleatoric uncertainty, the mixture model's uncertainty score consists of the weighted aleatoric and mixture components, and it is more effective in reflecting the prediction error extent as a function of the uncertainty.
Meanwhile, the mixture model has lower errors on all quantile-level intervals.
This also explains the overall competitive prediction performance of mixture models in Table~\ref{tab:baselines} from the uncertainty perspective.

\section{Conclusion}\label{sec:conclusion}
This paper focuses on probabilistic forecasting with multi-source time series data.
We present a neural mixture structure based probability model on multi-source time series and the associated phased learning method.
The experiment results exhibit the competitive performance of the proposed model and learning method on both point and probabilistic metrics. 
It reveals that capturing data-source-wise predictive relations and their adaptive combinations is important for exploiting the predictive power of multi-source data. 
Meanwhile, for mixture models with hierarchical and interacted components, understanding the learning behaviors of these components and accordingly customizing the learning method help realize the capacity of models. 

The limitation of the present work lies in the representation learning of time series data. 
It will be worth exploring more advanced representation learning methods and studying the effect on prediction performance.

\clearpage
\newpage
\bibliographystyle{IEEEtran}
\bibliography{reference} 

\onecolumn
\newpage
\begin{appendices}

\section{Model Specification}\label{appendix:model}

\subsection{Target Distributions}

We show the flexibility of specifying the prediction modules of the mixture models through two different target distributions. 

For the target variable in a normal distribution, i.e., $y_t \in \mathbb{R}$, the prediction module of each data source is expressed as:
\begin{align}
p_{\omega_s}(y_t \,|\, z_t = s, {\mathbf{h}}_{s}) = \frac{1}{\sqrt{2\pi} \sigma_{s,t}} \exp\left( -\frac{(y_t - \mu_{s,t})^2}{2\sigma^2_{s,t}} \right)
\end{align}
$\mu_{s,t}$ and $\sigma_{s,t}^2$ are derived by $g_{w_s}(\mathbf{h}_{s,t})$, where $g_{w_s}( \cdot )$ can be any flexible function such as dense layers parameterized by $\omega_s$~\cite{salinas2020deepar}.
The corresponding predictive mean and variance are $\mathbb{E}[{y}_{t} \,|\, z_t=s, \widetilde{\mathbf{h}}_s ] = \mu_{s,t}$ and $\text{Var}(y_t | z_t = s, \widetilde{\mathbf{h}}_{s,t}) = \sigma_{s,t}^2$.

When $y_t \in \mathbb{R}^+$ and is assumed to follow the log-normal distribution, the density function is:
\begin{align}
p_{\omega_s}(y_t \,|\, z_t = s, {\mathbf{h}}_{s}) = \frac{1}{\sqrt{2\pi} y_t \sigma_{s,t}} \exp\left( -\frac{(\ln y_t - \mu_{s,t})^2}{2\sigma^2_{s,t}} \right)
\end{align}
The predictive mean and variance by the data source $s$ are $\mathbb{E}[{y}_{t} \,|\, z_t=s, \widetilde{\mathbf{h}}_s ] = \exp(\mu_{s,t} + \frac{1}{2}\sigma_{s,t}^2)$ and $\text{Var}(y_t | z_t = s, \widetilde{\mathbf{h}}_{s,t}) = \exp(\sigma_{s,t}^2 -1) \cdot \exp(2 \mu_{s,t} + \sigma_{s,t}^2)$~\cite{crow1987lognormal, mackay2003information}.

\subsection{Inference}\label{appendix:infer}

In this part, we present several predictive quantities that can be derived from the mixture model. 

\textbf{Point Prediction.}
It is the predictive mean defined as:
\begin{align}\label{eq:mean_deri}
\hat{y}_t \triangleq \mathbb{E}[y_t \,|\, \mathbf{x}_{1:S, <t} ] 
& = 
\int \sum_{s=1}^{S} \mathbb{P}_{\theta}(z_{t} = s \,|\, \cdot ) p_{\omega_s}(y_t | z_t = s, \widetilde{\mathbf{h}}_{s}) \, y_t \, \text{d}y_t \\
& =
\sum_{s=1}^S \mathbb{P}_{\theta}(z_{t} = s \,|\, \cdot ) \mathbb{E}[{y}_{t} | z_t=s, \widetilde{\mathbf{h}}_{s,t}]
\end{align}
, where $\mathbb{E}[{y}_{t} \,|\, z_t=s, \widetilde{\mathbf{h}}_s ]$ denotes the predictive mean by source $s$. 
As aforementioned, it is $\mu_{s,t}$ for the normal distribution and $\exp(\mu_{s,t} + \frac{1}{2}\sigma_{s,t}^2)$ for the log-normal distribution~\cite{mackay2003information}.

\textbf{Uncertainty Score.}
According to the variance definition, we have:
\begin{align}\label{eq:var_deri}
\begin{split}
& \hat{u}_t \triangleq \text{Var}(\, {y}_{t} \,|\, \mathbf{x}_{1:S, <t} ) = \mathbb{E}[ y_t^2 \,|\, \mathbf{x}_{1:S, <t} ] - \mathbb{E}^2[y_t \,|\, \mathbf{x}_{1:S, <t}] 
\end{split}
\end{align}
Plugging the following into Eq.\ref{eq:var_deri} gives rise to Eq.\ref{eq:var}
\begin{align}
& \mathbb{E}[ y_t^2 \,|\, \mathbf{x}_{1:S, <t} ] = \sum_{s=1}^S \mathbb{P}_{\theta}(z_{t} = s \,|\, \cdot) \mathbb{E}[ y_t^2 | z_t =s, \widetilde{\mathbf{h}}_{s} ] 
\\
& \mathbb{E}[ y_t^2 | z_t =s, \widetilde{\mathbf{h}}_{s} ] = \text{Var}(y_t | z_t = s, \widetilde{\mathbf{h}}_{s,t}) + \mathbb{E}^2[y_t | z_t =s, \widetilde{\mathbf{h}}_{s} ]
\end{align}
, where the variance term $\text{Var}(y_t | z_t = s, \widetilde{\mathbf{h}}_{s,t})$ is $\sigma_{s,t}^2$ for the normal distribution, and $\exp(\sigma_{s,t}^2 -1) \cdot \exp(2 \gamma_{s,t} + \sigma_{s,t}^2)$ for the log-normal distribution~\cite{crow1987lognormal}.

\textbf{Quantile Prediction.}
Based on the mixture model, the predictive cumulative probability function (CDF) of the target is expressed as:
\begin{align}
\begin{split}
P(y_t \,|\, \mathbf{x}_{1:S,<t}) & = \int p(\, \mathbf{h}_{1:S,t} \,|\, \mathbf{x}_{1:S,<t} ) \cdot \sum_{s = 1}^{S} \mathbb{P}_{\theta}(z_t = s \,|\, \mathbf{h}_{1:S,t} )
\cdot P_{\omega_s}(y_t \,|\, z_t = s, \mathbf{h}_{s,t}) \, \text{d}\mathbf{h}_{1:S,t} 
\\
& = \sum_{s = 1}^{S} \mathbb{P}_{\theta}(z_t = s \,|\, \widetilde{\mathbf{h}}_{1:S,t} ) \cdot P_{\omega_s}(y_t \,|\, z_t = s, \widetilde{\mathbf{h}}_{s,t})
\end{split}
\end{align}
where $\widetilde{\mathbf{h}}_{s,t} = \text{NN}_{\eta_s}(\mathbf{x}_{s, <t}) $, and $P(y_t \,|\, z_t = s, \widetilde{\mathbf{h}}_{s,t})$ represents the individual data source's CDF. 
The second equality is due to the delta distribution applied to $p( \, \mathbf{h}_{1:S,t} \,|\, \mathbf{x}_{1:S,<t} )$, as is presented in Sec.\ref{sec:model_structure}.

Then, for the quantile level $\alpha \in [ 0, 1 ]$, we resort to a numerical method by applying the root-finding algorithm to the function in Eq.~\ref{eq:root_func_appendix}, while a sampling-based method can also be applied.
The solution of the root-finding algorithm is the quantile prediction $\hat{y}_{t,\alpha}$.
\begin{align}\label{eq:root_func_appendix}
Q\big( y \,; \alpha, P( \cdot \,|\, \mathbf{x}_{1:S,<t}) \big) \triangleq P( {y} \,|\, \mathbf{x}_{1:S,<t} ) - \alpha
\end{align}

\textbf{Prediction Interval.}
For obtaining a prediction interval corresponding to a probability range, we make use of the corresponding quantile predictions of the probability range.
For instance, a prediction interval corresponding to the probability range $[0.1, 0.9]$ is constructed by the root solutions of $ Q\big( y \,; \alpha = 0.1, P( \cdot \,|\, \mathbf{x}_{1:S,<t}) \big) = 0$ and $Q\big( y \,; \alpha = 0.9, P( \cdot \,|\, \mathbf{x}_{1:S,<t}) \big) = 0$.

\subsection{Learning}

The loss function of the mixture model is:
\begin{align}\label{eq:loss_share}
\begin{split}
\mathcal{L}\left( \Theta \,; \mathcal{D} \right) 
\triangleq 
\frac{1}{|\mathcal{T}|} \sum_{t \in \mathcal{T}}
-\log \mathbb{E}_{\mathbf{h}_{1:S,t}} 
\left[
\sum_{s = 1}^{S} \mathbb{P}_{\theta}(z_t = s \,|\, {\mathbf{h}}_{1:S,t}) \, p_{{\omega}_{s}}(y_t \,|\, z_t = s, {\mathbf{h}}_{s,t}) 
\right] 
\end{split}
\end{align}
where $\Theta = \big\{ \{\eta_s, \omega_s \}_{s=1}^S, \theta \big\}$ denotes the set of all trainable parameters.

Since in this paper the delta distribution is applied to $p( \, \mathbf{h}_{1:S,t} \,|\, \mathbf{x}_{1:S,<t} )$ and $p( \, \mathbf{r}_{1:S,t} \,|\, \mathbf{x}_{1:S,<t} )$, the integral over the hidden states is reduced to single points, i.e., 
$\widetilde{\mathbf{h}}_{s,t} = \text{NN}_{\eta_s}(\mathbf{x}_{s, <t})$. 

\section{Proofs}

\textbf{Lemma 4.1.}
\textit{
In minimizing the mixture loss via SGD optimization, given a random training sample indexed by $t$, the vectorized derivative of a source's prediction and representation modules are denoted by:
\begin{align}
& \mathbf{g}_{\omega_{s}, t} \triangleq -\frac{\partial \log p_{\mathbf{\omega}_{s}}(y_t \,|\, z_t = s, \widetilde{\mathbf{h}}_{s,t})}{\partial \omega_{s}} \Big|_{\Theta} 
\\
&\mathbf{g}_{\eta_{s}, t} \triangleq -\frac{\partial \log p_{\omega_{s}}(y_t \,|\, z_t = s, \widetilde{\mathbf{h}}_{s,t})}{\partial \widetilde{\mathbf{h}}_{s,t}} \nabla\widetilde{\mathbf{h}}_{s,t} \Big|_{\Theta} 
\\
& \mathbf{d}_{\eta_{s}, t} \triangleq -\sum_{k=1}^{S} \pi_{k,t} \frac{\partial \log\mathbb{P}_{\theta}(z_{t} = k | \cdot)}{\partial \widetilde{\mathbf{h}}_{s,t}} \nabla\widetilde{\mathbf{h}}_{s,t} \Big|_{\Theta} 
\end{align}
where $\nabla\widetilde{\mathbf{h}}_{s,t}$ denotes the Jacobian matrix of source $s$'s representation w.r.t. $\eta_s$.
Meanwhile, the posterior weight $\pi_{s, t}$ is defined as:
\begin{align}\label{eq:post_weight}
\begin{split}
\pi_{s,t} 
& \triangleq p(z_t = s \,|\, y_t, \mathbf{x}_{1:S,<t} ; \Theta)
\\
& = \frac{p_{{\omega}_{s}}(y_t \,|\, z_t = s, \widetilde{\mathbf{h}}_{s,t}) \, \mathbb{P}_{\theta}(z_t = s \,|\, \cdot)}{\sum_{k = 1}^{S} p_{{\omega}_{k}}(y_t \,|\, z_t = k, \widetilde{\mathbf{h}}_{k,t}) \, \mathbb{P}_{\theta}(z_t = k \,|\, \cdot)}  
\end{split}
\end{align}
Then, the (stochastic) updates to the prediction and representation parameters of source $s$, i.e., $\omega_{s}$ and $\eta_{s}$, are:
\vspace{-0.2cm}
\begin{align}\label{eq:update}
\pi_{s,t} \, \mathbf{g}_{\omega_{s},t} \,\,\,\text{and}\text{ }\,\,\, \pi_{s,t} \, \mathbf{g}_{\eta_{s},t} + \mathbf{d}_{\eta_{s},t}
\end{align}
}

\begin{proof}

Given a random training sample $(y_t, \mathbf{x}_{1:S,<t})$, the mixture loss Eq.\ref{eq:loss_share} is expressed: 
\begin{align*}
& -\log \sum_{k = 1}^{S} \mathbb{P}_{\theta}(z_t = k | \widetilde{\mathbf{h}}_{1:S,t}) \, p_{{\omega}_{k}}(y_t | z_t = k, \widetilde{\mathbf{h}}_{k,t})
\, \text{ where } \, \widetilde{\mathbf{h}}_{k,t} = \text{NN}_{\eta_k}(\mathbf{x}_{k, <t})
\end{align*}

Take the derivative w.r.t. the prediction module parameter of source $s$, 
\begin{align}
\begin{split}\label{eq:lemma1_1}
& \frac{\partial -\log \sum_{k = 1}^{S} \mathbb{P}_{\theta}(z_t = k | \widetilde{\mathbf{h}}_{1:S,t}) \, p_{{\omega}_{k}}(y_t | z_t = k,\widetilde{\mathbf{h}}_{k,t})}{\partial \omega_s}  \Big|_{\Theta} 
\\ 
& =
-\frac{\mathbb{P}_{\theta}(z_t = s | \widetilde{\mathbf{h}}_{1:S,t})}{\sum_{k = 1}^{S} \mathbb{P}_{\theta}(z_t = k | \widetilde{\mathbf{h}}_{1:S,t}) \, p_{{\omega}_{k}}(y_t | z_t = k,\widetilde{\mathbf{h}}_{k,t})} 
\frac{\partial p_{{\omega}_{s}}(y_t | z_t = s,\widetilde{\mathbf{h}}_{s,t})}{\partial \omega_s} \Big|_{\Theta} 
\\ 
& = 
-\underbrace{\frac{\mathbb{P}_{\theta}(z_t = s | \widetilde{\mathbf{h}}_{1:S,t}) p_{{\omega}_{s}}(y_t | z_t = s,\widetilde{\mathbf{h}}_{s,t}) }{\sum_{k = 1}^{S} \mathbb{P}_{\theta}(z_t = k | \widetilde{\mathbf{h}}_{1:S,t}) \, p_{{\omega}_{k}}(y_t | z_t = k,\widetilde{\mathbf{h}}_{k,t})}}_{\pi_{s,t}}
\frac{\partial \log p_{{\omega}_{s}}(y_t | z_t = s,\widetilde{\mathbf{h}}_{s,t})}{\partial \omega_s} \Big|_{\Theta} 
\end{split}
\end{align}

In Eq.\ref{eq:lemma1_1}, the second equality is through the derivative trick of log functions, i.e., $\partial f(\omega) / \partial \omega =  f(\omega) \, \partial \log f(\omega) / \partial \omega$.

Likewise, for $\eta_s$, we have the following:
\begin{align}
\begin{split}\label{eq:lemma1_2}
& \frac{\partial -\log \sum_{k = 1}^{S} \mathbb{P}_{\theta}(z_t = k | \widetilde{\mathbf{h}}_{1:S,t}) \, p_{{\omega}_{k}}(y_t | z_t = k,\widetilde{\mathbf{h}}_{k,t})}{\partial \eta_s} \Big|_{\Theta} 
\\ 
& =
-\frac{1}{\sum_{k = 1}^{S} \mathbb{P}_{\theta}(z_t = k | \widetilde{\mathbf{h}}_{1:S,t}) \, p_{{\omega}_{k}}(y_t | z_t = k,\widetilde{\mathbf{h}}_{k,t})} 
\sum_{k = 1}^{S} \frac{ \partial \, \mathbb{P}_{\theta}(z_t = k | \widetilde{\mathbf{h}}_{1:S,t}) p_{{\omega}_{k}}(y_t | z_t = k,\widetilde{\mathbf{h}}_{k,t})}{\partial \eta_s} \Big|_{\Theta} 
\\ 
& =
-\sum_{k = 1}^{S} \pi_{k,t} \frac{ \partial \log \mathbb{P}_{\theta}(z_t = k | \widetilde{\mathbf{h}}_{1:S,t}) p_{{\omega}_{k}}(y_t | z_t = k,\widetilde{\mathbf{h}}_{k,t}) }{ \partial \eta_s } \Big|_{\Theta}
\\ 
& =
-\sum_{k = 1}^{S} \pi_{k,t} \frac{ \partial \log \mathbb{P}_{\theta}(z_t = k | \widetilde{\mathbf{h}}_{1:S,t}) }{ \partial \eta_s } - \pi_{s,t} \frac{ \partial \log p_{{\omega}_{s}}(y_t | z_t = s,\widetilde{\mathbf{h}}_{s,t}) }{ \partial \eta_s } \Big|_{\Theta}
\end{split}
\end{align}

In Eq.\ref{eq:lemma1_2}, the second equality is based on the derivative trick of log functions.
Because the weight module $\mathbb{P}_{\theta}(z_t = k | \widetilde{\mathbf{h}}_{1:S,t})$ takes as input the representations of all data sources, there exists the sum of derivatives in the first term in the last equality.
Meanwhile, since $\widetilde{\mathbf{h}}_{1:S,t}$ is the output of representation modules parameterized by $\{\eta_s\}_{s=1}^S$, the derivative w.r.t. $\eta_s$ is through $\widetilde{\mathbf{h}}_{s,t}$. 
Then, the update scheme in Eq.~\ref{eq:update} is obtained. 

\end{proof}

\textbf{Lemma 4.2.}
\textit{
In the iterative minimization process of the mixture loss, at any point of $\Theta$, there exists the following upper bound:
\begin{align}\label{eq:bound}
\begin{split}
\mathcal{L}\left( \Theta \,; D_t\right) 
& \leq 
\overline{\mathcal{L}}\big( \{\eta_s, \omega_s\}_{s=1}^S \,; D_t \big)
\end{split}
\end{align}
where $D_t$ is a random training sample $D_t \triangleq ( y_t, \mathbf{x}_{1:S,<t} ) \in \mathcal{D}$.
\begin{align}
\begin{split}
\overline{\mathcal{L}}\big( \{\eta_s, \omega_s\}_{s=1}^S  \,;  D_t \big) \triangleq - \sum_{s = 1}^{S} \frac{1}{S} \log  p_{\omega_s}(y_t \,|\, z_t = s, \widetilde{\mathbf{h}}_{s,t} ) - \log S a_{t}^{*}
\end{split}
\end{align}
and $a_{t}^{*}$ is a constant value, $a_{t}^{*} = \min_{s}(\, \{ \, \mathbb{P}(z_t = s \,|\, \cdot) \} \,)$. 
}

\begin{proof}


The mixture loss on a random training sample $D_t$ is expressed:
\begin{align}
\mathcal{L}\left( \Theta \,; D_t \right)
\triangleq 
-\log \sum_{s = 1}^{S} \mathbb{P}_{\theta}(z_t = s \,|\, \widetilde{\mathbf{h}}_{1:S,t} ) \, p_{{\omega}_{s}}(y_t \,|\, z_t = s, \widetilde{\mathbf{h}}_{s,t})
\end{align}
where $\widetilde{\mathbf{h}}_{s,t} = \text{NN}_{\eta_s}(\mathbf{x}_{s, <t})$.

Given $\Theta$, by defining $a_{t}^{*} \triangleq \min_{s} \left(\, \{ \, \mathbb{P}(z_t = s \,|\, \cdot) \}_{s=1}^{S} \, \right)$, we have

\begin{align}\label{eq:ineql_alpha}
\begin{split}
\mathcal{L}\left( \Theta \,; D_t \right) 
\leq
-\log \sum_{s=1}^{S} a_{t}^{*} \, p_{{\omega}_{s}}(y_t \,|\, z_t = s, \widetilde{\mathbf{h}}_{s,t})
\end{split}
\end{align}
Eq.~\ref{eq:ineql_alpha} holds, because the probability densities are non-negative.
The equality holds when all weights are equal.

Then, Eq.~\ref{eq:ineql_alpha} is further expressed as follows:
\begin{align}\label{eq:ineql_jesen}
& = 
- \log \sum_{s=1}^{S} \frac{a_{t}^{*}}{S a_{t}^{*}} \, p_{{\omega}_{s}}(y_t \,|\, z_t = s, \widetilde{\mathbf{h}}_{s,t}) - \log S a_{t}^{*}
\\ 
& \leq 
- \sum_{s=1}^{S} \frac{1}{S} \log p_{{\omega}_{s}}(y_t \,|\, z_t = s, \widetilde{\mathbf{h}}_{s,t}) - \log S a_{t}^{*}
\end{align}
$S$ is the number of data sources.
The last inequality is based on Jensen's inequality.
\end{proof}

\textbf{Theorem 4.3.}
\textit{
For all stochastic derivatives $\mathbf{g}_{} \in \{ \mathbf{g}_{\omega_{s}}, \mathbf{g}_{\eta_{s}}, \mathbf{d}_{\eta_{s}} \}$ corresponding to the data source $s$, suppose it is bounded $\mathbb{E}\lVert \mathbf{g}_{} \rVert^2 \leq G^2$.
The trainable parameters related to a data source's prediction are denoted by 
$\Theta_{s} = [ \omega_{s}, \eta_{s} ]^\top$ and the corresponding gradient is  $\mathcal{G}_s = \mathbb{E}[\mathbf{g}_{\omega_{s}}, \mathbf{g}_{\eta_{s}}]^{\top}$.
Over the learning step $i = 0, \cdots, I$, the convergence of $\mathcal{G}_s$ to a stationary point in the direct optimization and the impartial-phase optimization are respectively: 
\begin{align}
\text{Direct:} \,\,\, & \frac{1}{I} \sum_{i=0}^{I-1} \mathbb{E} \norm{\mathcal{G}_{s,i}}^2 
\leq
  \sqrt{10} G \sqrt{ \frac{ L \left( \mathcal{L}_{s,0} - \mathcal{L}_{s,I} \right) }{I \pi_{s}^{*}} } 
  + \sum_{i=0}^{I-1} \frac{1}{2I} \left( \left\lVert \mathbb{E}[ \mathbf{g}_{\eta_s,i} ] - \mathbb{E}[ \mathbf{d}_{\eta_s,i} ] \right\rVert^2 \right)
\\
\text{Impartial phase:} \,\,\, & \frac{1}{I} \sum_{i=0}^{I-1} \mathbb{E} \norm{\mathcal{G}_{s,i}}^2 
\leq 
  2 G \sqrt{ \frac{ L \left(\mathcal{L}_{s, 0} - \mathcal{L}_{s, I} \right) }{I}}
\end{align}
}
, where $\pi_{s}^{*} \in (0, 1)$ is the minimum of poster weights across data samples and iteration steps.
$\mathcal{L}_{s, i}$ represents the negative log-likelihood of the predictions by data source $s$ at step $i$.

\begin{proof}

Before going to the proof details, we first recall some notations. 

As defined in {Lemma 4.1.}, we have the (stochastic) derivatives on a random training sample $t$ as follows:

\begin{align*}
& \mathbf{g}_{\omega_{s} } \triangleq -\frac{\partial \log p_{\mathbf{\omega}_{s}}(y_t \,|\, z_t = s, \widetilde{\mathbf{h}}_{s,t})}{\partial \omega_{s}} \Big|_{\Theta_i} 
\\
&\mathbf{g}_{\eta_{s} } \triangleq -\frac{\partial \log p_{\omega_{s}}(y_t \,|\, z_t = s, \widetilde{\mathbf{h}}_{s,t})}{\partial \widetilde{\mathbf{h}}_{s, t}} \nabla\widetilde{\mathbf{h}}_{s, t} \Big|_{\Theta_i} 
\\
& \mathbf{d}_{\eta_{s} } \triangleq -\sum_{k=1}^{S} \pi_{k,i} \frac{\partial \log\mathbb{P}_{\theta}(z_{t} = k | \cdot)}{\partial \widetilde{\mathbf{h}}_{s,t}} \nabla\widetilde{\mathbf{h}}_{s, t} \Big|_{\Theta_i}
\end{align*}
where $\nabla\widetilde{\mathbf{h}}_{s,t}$ denotes the Jacobian matrix of source $s$'s representation module w.r.t. $\eta_s$.

Given the parameter $\Theta_{s, i}$ at step $i$, the predictive performance of data source $s$ is reflected in the negative log-likelihood which is a function of $\Theta_{s, i}$ as:
\begin{align}
\mathcal{L}_{s, i} \triangleq \frac{1}{|\mathcal{T}|} \sum_{t \in \mathcal{T}} 
- \log \mathbb{E}_{\mathbf{h}_{s,t}} 
  \left[ p_{{\omega}_{s}}(y_t \,|\, z_t = s, {\mathbf{h}}_{s,t}) \right]
\end{align}

Then, the empirical expected gradient of $\Theta_{s, i}$ w.r.t. $\mathcal{L}_{s, i}$ is as follows: (the step-index $i$ is omitted for simplicity)
\begin{align}
\mathcal{G}_{s} \triangleq \mathbb{E}[\mathbf{g}_{\omega_{s}}, \mathbf{g}_{\eta_{s}}]^{\top} 
                = \frac{1}{|\mathcal{T}|} \sum_{t \in \mathcal{T}} [\mathbf{g}_{\omega_{s},t}, \mathbf{g}_{\eta_{s},t}]^{\top}
\end{align}

\underline{Direct optimization}

In the direct optimization, as shown in {Lemma 4.1.}, the stochastic update to the prediction and representation parameters of a data sources $s$ is expressed as $\widehat{\mathbf{g}}_{s,i} \triangleq [ \pi_{s,i} \, \mathbf{g}_{\omega_{s}, i} ,\, \pi_{s,i} \, \mathbf{g}_{\eta_{s}, i} + \mathbf{d}_{\eta_{s}, i} ]^{\top}$, where $\pi_{s,i}$ is the posterior weight.
In the following, we will drop the random training sample subscript $t$ for simplicity when unnecessary.

Following the $L$-smoothness assumption of $\mathcal{L}_{s}$ yields:
\begin{align}
\mathcal{L}_{s, i+1} - \mathcal{L}_{s, i} 
& \leq 
  \left\langle \mathcal{G}_{s,i}, - \gamma \mathbb{E}[ \widehat{\mathbf{g}}_{s,i} ] \right\rangle
  + \frac{L}{2} \mathbb{E}\left[ \left\langle -\gamma \widehat{\mathbf{g}}_{s,i} ,  -\gamma \widehat{\mathbf{g}}_{s,i} \right\rangle \right] 
\\
& =
  - \gamma \left\langle \mathbb{E}\begin{bmatrix} \mathbf{g}_{\omega_{s,i}} \\ \mathbf{g}_{\eta_{s,i}} \end{bmatrix} , 
                        \mathbb{E}\begin{bmatrix} \pi_{s,i} \, \mathbf{g}_{\omega_{s,i}} \\ \pi_{s,i} \, \mathbf{g}_{\eta_{s,i}} \end{bmatrix} \right\rangle
  - \gamma \left\langle \mathbb{E}[ \mathbf{g}_{\eta_{s,i}} ] , \mathbb{E}[\mathbf{d}_{\eta_{s,i}}] \right\rangle
  + \frac{L}{2} \mathbb{E}\left[ \left\langle -\gamma \widehat{\mathbf{g}}_{s,i} ,  -\gamma \widehat{\mathbf{g}}_{s,i} \right\rangle \right] 
\\
& \leq 
  - \gamma \, \pi_{s,i}^{*}  \left\lVert \mathbb{E}\begin{bmatrix} \mathbf{g}_{\omega_{s,i}} \\ \mathbf{g}_{\eta_{s,i}} \end{bmatrix} \right\rVert^2
  \underbrace{- \gamma \left\langle \mathbb{E}[ \mathbf{g}_{\eta_{s,i}} ] , \mathbb{E}[\mathbf{d}_{\eta_{s,i}}] \right\rangle}_{(a)}
  + \underbrace{\frac{L}{2} \mathbb{E}\left[ \left\langle -\gamma \widehat{\mathbf{g}}_{s,i} ,  -\gamma \widehat{\mathbf{g}}_{s,i} \right\rangle \right]}_{(b)} \label{eq:theory_5}
\end{align}
where $\gamma$ is the learning step size.
$\pi_{s,i}^{*}$ is the minimum of posterior weights across data samples, i.e., $\pi_{s,i}^{*} = \underset{ t }{\min} ( \{ \pi_{s, t,i} \}_t )$, and it is always less than $1$, i.e., $\pi_{s, i}^{*} \in (0, 1)$.
$\langle \cdot \, , \cdot \rangle$ is the  inner product of two vectors.

Then, for the $(a)$ and $(b)$ terms, we have the following:
\begin{align}
(a)
& = - \frac{\gamma}{2} \left\lVert \mathbb{E}[ \mathbf{g}_{\eta_s,i} ] \right\rVert^2
    - \frac{\gamma}{2} \left\lVert \mathbb{E}[ \mathbf{d}_{\eta_s,i} ] \right\rVert^2
    + \frac{\gamma}{2} \left\lVert \mathbb{E}[ \mathbf{g}_{\eta_s,i} ] - \mathbb{E}[ \mathbf{d}_{\eta_s,i} ] \right\rVert^2 
\\
& \leq 
    \frac{\gamma}{2} \left\lVert \mathbb{E}[ \mathbf{g}_{\eta_s,i} ] - \mathbb{E}[ \mathbf{d}_{\eta_s,i} ] \right\rVert^2 \label{eq:theory_3} 
\end{align}

\begin{align}
(b)
& = 
  \frac{L \gamma^2}{2} \mathbb{E} \left\lVert\begin{bmatrix} \pi_{s,i} \, \mathbf{g}_{\omega_{s,i}} \\ \pi_{s,i} \, \mathbf{g}_{\eta_{s,i}} 
  + \mathbf{d}_{\eta_{s,i}} \end{bmatrix} \right\rVert^2 \\
& =
  \frac{L \gamma^2}{2} \mathbb{E} \left[
  \pi_{s,i}^2 \norm{\mathbf{g}_{\omega_{s,i}}}^2 + \pi_{s,i}^2 \norm{\mathbf{g}_{\omega_{s,i}}}^2 + 2 \pi_{s,i} \langle \mathbf{g}_{\eta_{s,i}} \, , \mathbf{d}_{\eta_{s,i}} \rangle + \norm{\mathbf{d}_{\eta_{s,i}}}^2
  \right] \\
& \leq
  \frac{L \gamma^2}{2} \mathbb{E} \left[
  \pi_{s,i}^2 \norm{\mathbf{g}_{\omega_{s,i}}}^2 + \pi_{s,i}^2 \norm{\mathbf{g}_{\omega_{s,i}}}^2 
  + 
  \pi_{s,i} \norm{\mathbf{g}_{\eta_{s,i}}}^2 + \pi_{s,i} \norm{\mathbf{d}_{\eta_{s,i}}}^2 + \norm{\mathbf{d}_{\eta_{s,i}}}^2
  \right] \\
& \leq
  \frac{5}{2} L \gamma^2 G^2 \label{eq:theory_4} 
\end{align}
The first inequality is because $ 2\langle \mathbf{a}, \mathbf{b} \rangle \leq \norm{ \mathbf{a} }^2 + \norm{ \mathbf{b} }^2$ and the last one is due to the gradient bound assumption.

Plugging Eq.\ref{eq:theory_3} and \ref{eq:theory_4} into Eq.\ref{eq:theory_5} and summing the inequality from $i = 0$ to $i = I$ lead to: 
\begin{align}
\mathcal{L}_{s, i+1} - \mathcal{L}_{s, i} 
& \leq 
  - \gamma \sum_{i=0}^{I-1} \pi^{*}_{i} \big\lVert \mathcal{G}_{s,i} \big\rVert^2
  + \sum_{i=0}^{I-1} \frac{\gamma}{2} \left\lVert \mathbb{E}[ \mathbf{g}_{\eta_s,i} ] - \mathbb{E}[ \mathbf{d}_{\eta_s,i} ] \right\rVert^2
  + \frac{5}{2} L \gamma^2 G^2 I  \label{eq:theory_6}
\end{align}

If we further take the minimum of poster weights across iteration steps by setting $\pi^{*}_s = \underset{ i }{\min} (\{ \pi_{s, i}^{*} \}_{i})$ or equivalently 
$\pi^{*}_s = \underset{ t,i }{\min} (\{ \pi_{s,t,i} \}_{t,i})$, rearranging Eq.~\ref{eq:theory_6} gives:
\begin{align}
\frac{1}{I} \sum_{i=0}^{I-1} \mathbb{E} \norm{\mathcal{G}_{s,i}}^2
\leq
  \frac{\left( \mathcal{L}_{s,0} - \mathcal{L}_{s,I} \right) }{\gamma I \pi_{s}^{*}} 
  + \sum_{i=0}^{I-1} \frac{1}{2} \left( \left\lVert \mathbb{E}[ \mathbf{g}_{\eta_s,i} ] - \mathbb{E}[ \mathbf{d}_{\eta_s,i} ] \right\rVert^2 \right)
  + \frac{5}{2} L \gamma G^2  \label{eq:theory_7} 
\end{align}

Plugging in the step size $\gamma$ that minimizes the RHS of Eq.~\ref{eq:theory_7} shows the result of the direct optimization:
\begin{align}
\frac{1}{I} \sum_{i=0}^{I-1} \mathbb{E} \norm{\mathcal{G}_{s,i}}^2 
\leq
  \sqrt{10} G \sqrt{ \frac{ L \left( \mathcal{L}_{s,0} - \mathcal{L}_{s,I} \right) }{I \pi_{s}^{*}} } 
  + \sum_{i=0}^{I-1} \frac{1}{2I} \left( \left\lVert \mathbb{E}[ \mathbf{g}_{\eta_s,i} ] - \mathbb{E}[ \mathbf{d}_{\eta_s,i} ] \right\rVert^2 \right)
\end{align}

\underline{Impartial phase optimization}

In the impartial phase, the stochastic update to the prediction and representation parameters of a data source $s$ is expressed as: $\mathring{\mathbf{g}}_{s,i} \triangleq [ \frac{1}{S} \mathbf{g}_{\omega_{s}, i} ,\,  \frac{1}{S} \mathbf{g}_{\eta_{s}, i} ]^{\top}$.

Following the $L$-smoothness assumption yields:
\begin{align}
\mathcal{L}_{s, i+1} - \mathcal{L}_{s, i}
& \leq 
  \left\langle \mathcal{G}_{s, i} \, , - \gamma \mathbb{E}[ \mathring{\mathbf{g}}_{s,i} ] \right\rangle
  +
  \frac{L}{2} \mathbb{E} \left[ \left\langle -\gamma \mathring{\mathbf{g}}_{s,i} \, ,  -\gamma \mathring{\mathbf{g}}_{s,i} \right\rangle \right]
\\
& =
  -\frac{\gamma}{S} \norm{ \mathcal{G}_{s,i} }^2 + \frac{L \gamma^2}{2 S^2} \mathbb{E} \left[ \norm{\mathbf{g}_{\omega_s,i}}^2 + \norm{\mathbf{g}_{\eta_{s},i}}^2 \right]
\\
& \leq
  -\frac{\gamma}{S} \norm{ \mathcal{G}_{s,i} }^2 + \frac{L \gamma^2 G^2}{S^2} \label{eq:theory_1}
\end{align}

Then, by summing the inequality Eq.~\ref{eq:theory_1} from $i = 0$ to $i = I$ and some rearrangements, we obtain 
\begin{align}
\frac{1}{I} \sum_{i=0}^{I-1} \mathbb{E} \norm{\mathcal{G}_{s,i}}^2 
\leq 
   \frac{S \left( \mathcal{L}_{s, 0} - \mathcal{L}_{s, I} \right) }{\gamma I} + \frac{L \gamma G^2 }{S} \label{eq:theory_2}
\end{align} 

Plugging in the step size $\gamma$ that minimizes the RHS of Eq.~\ref{eq:theory_2} gives the convergence:
\begin{align}
\frac{1}{I} \sum_{i=0}^{I-1} \mathbb{E} \norm{\mathcal{G}_{s,i}}^2 
\leq 
   2 G \sqrt{ \frac{ L \left( \mathcal{L}_{s, 0} - \mathcal{L}_{s, I} \right) }{I}}
\end{align}

\end{proof}

\section{Additional Experiment Details and Results}\label{appendix:exp}

\textbf{Hyper-Parameters.}
The hyper-parameter search and associated training processes are run on a server with NVIDIA A100 GPUs.
For all models, Bayesian optimization is applied to search in the hyper-parameter space~\cite{bergstra2013making}.
Once the best hyperparameters are fixed, each model is retrained for five times with different random seeds, and the average performance is reported. 
For recurrent neural networks used in DAR, DF, and mixture models, the layer size and the number of layers are searched over $\{64, 128, 256, 512, 768, 1024\}$ and $\{1, 2, 3\}$.
The learning rate is chosen in the range $\{1e-5, 5e-5, 1e-4, 5e-4, 1e-3, 5e-3\}$ following an exponential decay scheme with a rate of $0.85$ every $10$ epochs. 
The search spaces of batch sizes, weight decay, and dropout rates respectively lie in $[256, 512]$, $\{1e-7, 5e-7, 1e-6, 5e-6, 1e-5, 5e-5, 1e-4,\}$, and $\{0.0, 0.1, 0.2, 0.3\}$

\textbf{Performance Metrics.}
Given the point prediction $\hat{y}_t$ and the true value $y_t$, the point prediction metrics are defined as follows:
\begin{equation}
\text{RMSE}=\sqrt{\frac{1}{ |\mathcal{T}_{\text{test}}| } \sum_{t \in \mathcal{T}_{\text{test}}}^{ }(y_t-\hat{y}_t)^2} 
\,\, \text{ and } \,\,
\text{MAE}=\frac{1}{ |\mathcal{T}_{\text{test}}| } \sum_{t \in \mathcal{T}_{\text{test}} }^{ } |y_t-\hat{y}_t|
\end{equation}
where $\mathcal{T}_{\text{test}}$ represents the set of timesteps in the testing data, and $|\mathcal{T}_{\text{test}}|$ is the number of testing examples.

As for the mean quantile loss (QLm), we define the (normalized) quantile loss of a given quantile level $\alpha \in (0, 1)$ as:
\begin{equation}
\text{QL}(\alpha)
=
  \frac{\sum_{t \in \mathcal{T}_{\text{test}}} 2 \left[ \alpha (y_t - \hat{y}_{t,\alpha}) \, \mathbb{I}_{y_t - \hat{y}_{t,\alpha} > 0} 
  + 
  (1-\alpha) (\hat{y}_{t,\alpha} - y_t) \, \mathbb{I}_{y_t - \hat{y}_{t,\alpha} \leq 0} \right] }{ \sum_{ t \in \mathcal{T}_{\text{test}} } | y_t | }
\end{equation}
where $\mathbb{I}$ takes value $1$ only when the corresponding condition holds~\cite{wang2019deep}.
$\hat{y}_{t,\alpha}$ is the quantile prediction obtained through the method presented in Sec.~\ref{appendix:infer}.

Then, the QLm is the mean value of the quantile losses $\text{QL}(\alpha)$ for $\alpha = 0.1, 0.3, 0.5, 0.7, 0.9$.

\textbf{Results.}
In this part, we report the performance result on all datasets.

Table~\ref{tab:point_metric_append} and \ref{tab:prob_metric_append} show the point and probabilistic prediction performance on all datasets.
Our mixture model \textbf{MIX} shows competitive performance over baselines.

\begin{table*}[!htbp]
\caption{
Results of point prediction metrics (mean and standard error respectively in the first and second line of each row).
The best result is marked by the grey box.
}
\centering
\resizebox{0.9\textwidth}{!}{
\begin{tabular}{l|llll|llll}
\toprule 
               & \multicolumn{4}{c|}{RMSE $\downarrow$}                           & \multicolumn{4}{c}{MAE $\downarrow$}\\
    Dataset    & DAR & DF & AF & \textbf{MIX}              & DAR & DF & AF & \textbf{MIX} \\
    \midrule
    \midrule
    Site 0     & 23.650 & 23.792 & 68.293   & \best{19.022}      & 12.015 & 13.743 & 35.196   & \best{10.297}  \\
               & 0.645  & 0.756  & 0.958    & 0.054              & 0.223  & 0.291  & 0.307    &  0.048          \\
    \midrule
    Site 1     & 20.012 & 31.467 & 66.199   & \best{18.517}      & 10.596 & 13.247 & 36.188   & \best{9.929}  \\
               & 0.205  & 0.570  & 0.912    & 0.086              & 0.109  & 0.772  & 0.462    &  0.027         \\
    \midrule
    Site 2     & 18.060 & 19.801 & 69.008   & \best{17.081}      & 9.054  & 11.007 & 53.221   & \best{8.573}  \\
               & 0.210  & 0.277  & 0.323    & 0.087              & 0.028  & 0.389  & 0.862    & 0.037         \\
    \midrule
    Site 3     & 29.134 & 40.351 & 95.691   & \best{21.969}      & 11.352 & 14.858 & 70.882   & \best{11.293}  \\
               & 0.147  & 0.574  & 0.823    & 0.101              & 0.108  & 0.229  & 0.677    &  0.059         \\
    \midrule
    Site 4     & 22.393 & 22.594 & 87.732   & \best{19.637}       & 10.883 & 14.151 & 66.508   & \best{10.638} \\
               & 0.278  & 0.292  & 0.932    &  0.062              & 0.013  & 0.505  & 0.709    & 0.084         \\
    \midrule
    Site 5     & 26.133 & 29.102 & 93.721   & \best{20.578}      & 11.341 & 11.234 & 66.004   & \best{10.318}  \\
               & 1.392  & 1.101  & 1.236    & 0.128              & 0.106  & 0.102  & 0.979    & 0.049          \\
    \midrule
    \midrule
    Exc. A    & 1.206   & 1.157   & 1.892    & \best{1.149}      & 0.925   & 0.916   & 1.021    & \best{0.898}  \\
              & 0.00342 & 0.00389 & 0.00253  & 0.00168           & 0.00276 & 0.00259 & 0.00324  & 0.00588      \\
    \midrule
    Exc. B    & 1.382  & 1.271   & 2.010   & \best{1.265}      & 0.995   & 0.981   & 1.032    & \best{0.977} \\
              & 0.0075 & 0.00177 & 0.0098  & 0.00302           & 0.00464 & 0.00101 & 0.00433  & 0.00197      \\
    \bottomrule
\end{tabular}
}
\label{tab:point_metric_append}
\end{table*}

\begin{table*}[!htbp]
\caption{
Results of probabilistic prediction metrics (mean and standard error respectively in the first and second line of each row).
The best result is marked by the grey box.
}
\centering
\resizebox{0.9\textwidth}{!}{
\begin{tabular}{l|llll|llll}
\toprule 
               & \multicolumn{4}{c|}{NLLm $\downarrow$}              & \multicolumn{4}{c}{QLm $\downarrow$}\\
    Dataset    & DAR & DF & AF & \textbf{MIX}     & DAR & DF & AF & \textbf{MIX} \\
    \midrule
    \midrule
    Site 0     & 3.922  & 4.196  & 5.439  & \best{3.879}          & 0.298  & 0.327  & 0.233  & \best{0.0463} \\
               & 0.011  & 0.013  & 0.004  & 0.011                 & 0.004  & 0.003  & 0.001  & 0.001         \\
    \midrule
    Site 1     & 3.860  & 3.992   & 5.208 & \best{3.696}         & 0.231   & 0.202   & 0.237 & \best{0.0505}  \\
               & 0.021  & 0.029   & 0.002 & 0.003                & 0.004   & 0.004   & 0.001 & 0.001          \\
    \midrule
    Site 2     & 3.745   & 3.945  & 5.124 & \best{3.679}       & 0.234   & 0.130   & 0.241 & \best{0.0502} \\
               & 0.005   & 0.014  & 0.001 & 0.001              & 0.002   & 0.008   & 0.001 & 0.001         \\
    \midrule
    Site 3     & 4.034  & 4.150   & 5.510 & \best{3.962}       & 0.277  & 0.199   & 0.233 & \best{0.0471}  \\
               & 0.016  & 0.015   & 0.001 & 0.001              & 0.011  & 0.007   & 0.001 & 0.001           \\
    \midrule
    Site 4     & 3.945  & 4.154   & 5.466 & \best{3.906}         & 0.276   & 0.191   & 0.227 & \best{0.0453}  \\
               & 0.016  & 0.007   & 0.003 & 0.005                & 0.006   & 0.009   & 0.001 & 0.003          \\
    \midrule
    Site 5     & 3.917 & 4.081   & 5.460 & \best{3.866}          & 0.274   & 0.533 & 0.235 & \best{0.0445}  \\
               & 0.015 & 0.009   & 0.001 & 0.001                 & 0.003   & 0.103 & 0.001 & 0.001          \\
    \midrule
    \midrule
    Exc. A    & 1.594   & 1.578   & 1.601     & \best{1.541}      & 2.974  & 4.498  & 4.982   & \best{2.840} \\
              & 0.00485 & 0.00550 & 0.00324   & 0.00258           & 0.0413 & 0.0694 & 0.0983  &  0.0624      \\
    \midrule
    Exc. B    & 1.698   & 1.660   & 1.723     & \best{1.647}      & 2.606  & 4.801 & 5.012   & \best{2.353} \\
              & 0.00853 & 0.00131 & 0.00732   & 0.00220           & 0.0958 & 0.119 & 0.1012  & 0.0930       \\
    \bottomrule
\end{tabular}
}
\label{tab:prob_metric_append}
\end{table*}

Table~\ref{tab:train_method_perf_append} shows the performance of the mixture models trained by the direct and phased learning methods.
The phased learning constantly helps enhance the performance.
This implies that when the prediction and representation related modules of mixture models are insufficiently learned yet, simultaneously learning the prediction modules with the weight modules may fail to reliably capture the predictive power of different data sources, thereby leading to less accurate weighted combinations.

\begin{table}[ht]
\centering
\caption{
Performance of the mixture models respectively trained by the direct and phased learning methods (mean and standard error respectively in the first and second line of each row).
}
      \begin{tabular}{l|ll|ll|ll|ll}
       \toprule       
                  & \multicolumn{2}{c|}{RMSE $\downarrow$} & \multicolumn{2}{c}{MAE $\downarrow$} & \multicolumn{2}{c|}{NLLm $\downarrow$} & \multicolumn{2}{c}{QLm $\downarrow$} \\
       Dataset    & Direct & Phased                        & Direct & Phased                      & Direct & Phased                        & Direct & Phased       \\
       \midrule
       \midrule
       Site 0  & 27.173 & \best{19.022}         & 14.905 & \best{10.297}    & 4.103 & \best{3.879}          & 0.0582 & \best{0.0463} \\
               & 1.617  & 0.054                 & 0.423  & 0.048            & 0.133 & 0.011                 & 0.009  & 0.001\\
       \midrule
       Site 1  & 41.549 & \best{18.517}         & 17.876 &\best{9.929}      & 3.772 & \best{3.696}          & 0.0589 & \best{0.0505}\\
               & 2.013  & 0.086                 & 1.102  & 0.027            & 0.006 & 0.003                 & 0.009  & 0.001 \\
       \midrule
       Site 2  & 17.279 & \best{17.081}         & 8.832 & \best{8.573}      & 3.720 & \best{3.679}          & 0.0550 & \best{0.0502}\\
               & 0.021  & 0.087                 & 0.011 & 0.037             & 0.013 & 0.001                 & 0.002  & 0.001 \\
       \midrule
       Site 3  & 30.509 & \best{21.969}         & 14.241 & \best{11.293}    & 4.392 & \best{3.962}          & 0.0669 & \best{0.0471}\\
               & 1.132  & 0.101                 & 0.102  & 0.059            & 0.029 & 0.001                 & 0.002  & 0.001\\
       \midrule
       Site 4  & 26.592 & \best{19.637}         & 17.908 & \best{10.638}    & 3.948 & \best{3.906}          & 0.0456 & \best{0.0453}\\
               & 1.123  & 0.062                 & 0.812  & 0.084            & 0.040 & 0.005                 & 0.003  & 0.003\\
       \midrule
       Site 5  & 30.233 & \best{20.578}         & 19.843 & \best{10.318}    & 3.891 & \best{3.866}          & 0.0448 & \best{0.0445} \\
               & 1.932  & 0.128                 & 1.324  & 0.049            & 0.013 & 0.001                 & 0.001  & 0.001 \\
       \midrule        
       \midrule
       Exc. A  & 1.172   & \best{1.149}         & 0.934  & \best{0.898}     & 1.596   & \best{1.541}        & 3.012  & \best{2.840} \\
               & 0.00123 & 0.00168              & 0.0356 & 0.00588          & 0.00216 & 0.00258             & 0.0563 & 0.0624 \\
       \midrule
       Exc. B  & 1.282   & \best{1.265}         & 0.998   & \best{0.977}    & 1.656   & \best{1.647}       & 2.510 & \best{2.353} \\
               & 0.00203 & 0.00302              & 0.00123 & 0.00197         & 0.00158 &  0.00220           & 0.0890 & 0.0930\\
       \bottomrule
       \end{tabular}
\label{tab:train_method_perf_append}
\end{table}

\begin{figure*}[!htbp]
\centering
\includegraphics[width=0.99\textwidth]{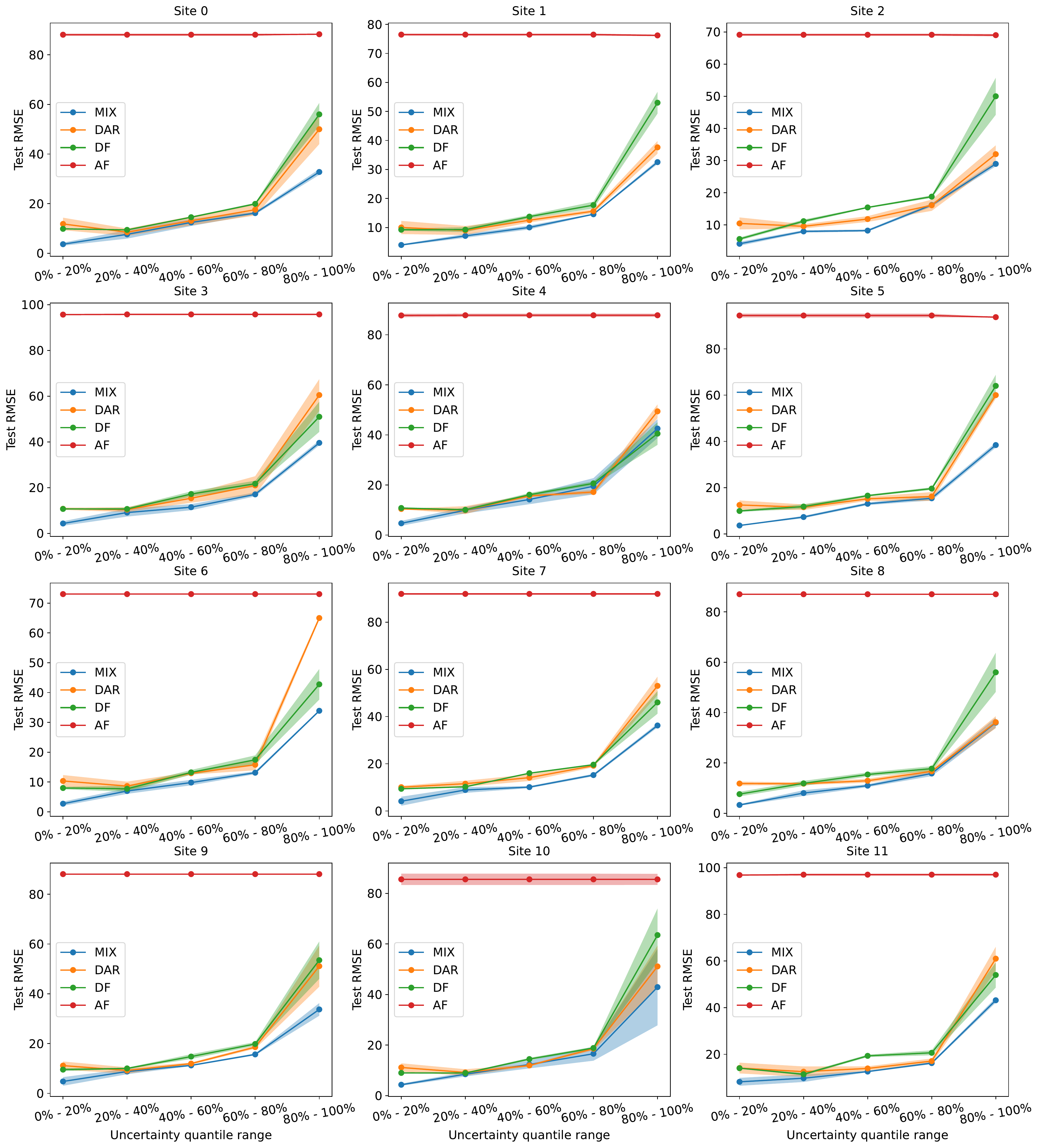}
\caption{Uncertainty Conditioned Prediction Errors.}
\label{fig:unc_error_append}
\end{figure*}

Fig.~\ref{fig:unc_error_append} show the uncertainty-conditioned prediction errors of all datasets.
The uncertainty quantile level on the x-axis of each sub-figure is derived from the empirical distribution of predictive uncertainties. 

\end{appendices}

\end{document}